\RequirePackage{fix-cm}
\PassOptionsToPackage{table}{xcolor}   %
\PassOptionsToPackage{hyphens}{url}
\documentclass{fairastra}

\usepackage[T1]{fontenc}
\usepackage[utf8]{inputenc}
\usepackage{lmodern}
\usepackage{newtxtext}
\usepackage{inconsolata}   %
\usepackage{microtype}
\usepackage{amssymb}
\usepackage{colortbl}
\usepackage{tabularx}      %
\usepackage[normalem]{ulem}
\usepackage{url}
\usepackage{eccvabbrv}

\newcommand{\gz}[1]{}

\def\reviewmode{1}
\ifnum\reviewmode=0
  \DeclareRobustCommand{\del}[1]{#1}
  \DeclareRobustCommand{\add}[1]{}
\else\ifnum\reviewmode=1
  \DeclareRobustCommand{\del}[1]{}
  \DeclareRobustCommand{\add}[1]{#1}
\else
  \DeclareRobustCommand{\del}[1]{\textcolor{gray}{\sout{#1}}}
  \DeclareRobustCommand{\add}[1]{\textcolor{red}{#1}}
\fi\fi
\DeclareRobustCommand{\yadd}[1]{\ifnum\reviewmode>0\ifnum\reviewmode=2\textcolor{yellow!70!black}{#1}\else#1\fi\fi}
\DeclareRobustCommand{\bdel}[1]{\textcolor{gray}{\sout{#1}}}
\DeclareRobustCommand{\badd}[1]{\textcolor{red}{#1}}

\title{Embodied Agents Take Control: Minimal-Interface Zero-Shot Agents
Rival Industrial-Scale Policies in Vision-and-Language Navigation}

\author{\parbox{\linewidth}{\raggedright
Jian Zhou\textsuperscript{1,*}\quad
Xunyi Zhao\textsuperscript{1,2,*}\quad
Gengze Zhou\textsuperscript{1}\quad
Zerui Li\textsuperscript{1}\quad
Sihao Lin\textsuperscript{1,2}\\[1pt]
Jiajun Liu\textsuperscript{2,3,4}\quad
Qi Wu\textsuperscript{1,2}
}}
\renewcommand{\affiliationlist}{{\small\sffamily\bfseries
\mbox{\textsuperscript{1}Australian Institute for Machine Learning, Adelaide University}\quad
\mbox{\textsuperscript{2}Responsible AI Research Centre}\quad
\mbox{\textsuperscript{3}CSIRO Data61}\quad
\mbox{\textsuperscript{4}The University of Queensland}\par}}
\contribution[*]{Equal contribution}
\abstract{Autonomous embodied agents must sustain a long decision-making loop of perceiving, acting, verifying, and self-correcting. In trained end-to-end policies and hand-designed workflows, the interaction structure remains externally specified. We study how much of this loop a general-purpose model can direct through a minimal interface. We term this organization \emph{agentic embodied control}: the reasoning model directly steers every action, keeping reasoning and control aligned. We evaluate fourteen frontier multimodal models as provided by their vendors, with monocular RGB and primitive actions across six navigation and embodied question-answering benchmarks. On 100-episode subsets, the best agents exceed published scores and approach or surpass our human tester. Who directs the loop matters: SmartWay and O2C-Nav using the same strong models achieve substantially lower success rates on R2R-CE. The lightweight mini-swe-agent harness performs comparably to vendor harnesses. Stronger models benefit from optional rather than imposed waypoints; adding SLAM raises R2R-CE success to 88\%. Physical trials expose limitations in body awareness, execution feedback, and persistent spatial understanding.}
\astradata[Project Page]{\url{https://jianzhou0420.github.io/mip/}}
\astradata[GitHub]{\url{https://github.com/jianzhou0420/MIP}}
\hypersetup{pdftitle={Embodied Agents Take Control: Minimal-Interface Zero-Shot Agents Rival Industrial-Scale Policies in Vision-and-Language Navigation},
  pdfauthor={Jian Zhou, Xunyi Zhao, Gengze Zhou, Zerui Li, Sihao Lin, Jiajun Liu, Qi Wu}}

\newcommand{\CsFails}{18}
\newcommand{\CsRunSR}{82}

\newcommand{\CsClaimN}{1}
\newcommand{\CsGambleN}{15}
\newcommand{\CsBeliefFailN}{2}
\newcommand{\CsOsrN}{5}
\newcommand{\CsNeMin}{3.0}
\newcommand{\CsNeMax}{34.1}

\newcommand{\CsRefinedN}{16}
\newcommand{\CsRevisedN}{2}

\newcommand{\CsDoubtActedN}{3}
\newcommand{\CsBlockedN}{10}
\newcommand{\CsAltVisibleN}{14}
\newcommand{\CsTopSceneN}{6}
\newcommand{\CsTopThreeScenesN}{11}

\newcommand{\CsUnderHundredN}{7}
\newcommand{\CsAN}{7}
\newcommand{\CsBN}{5}
\newcommand{\CsCN}{4}
\newcommand{\CsDN}{2}

\newcommand{\CsGreedyN}{11}

\newcommand{\CsStableN}{9}

\begin{document}
\noindent\makebox[\linewidth][s]{%
\makebox[0.31\linewidth][c]{\raisebox{-0.5\height}{\includegraphics[height=1.55cm,viewport=36 359 1882 1081,clip]{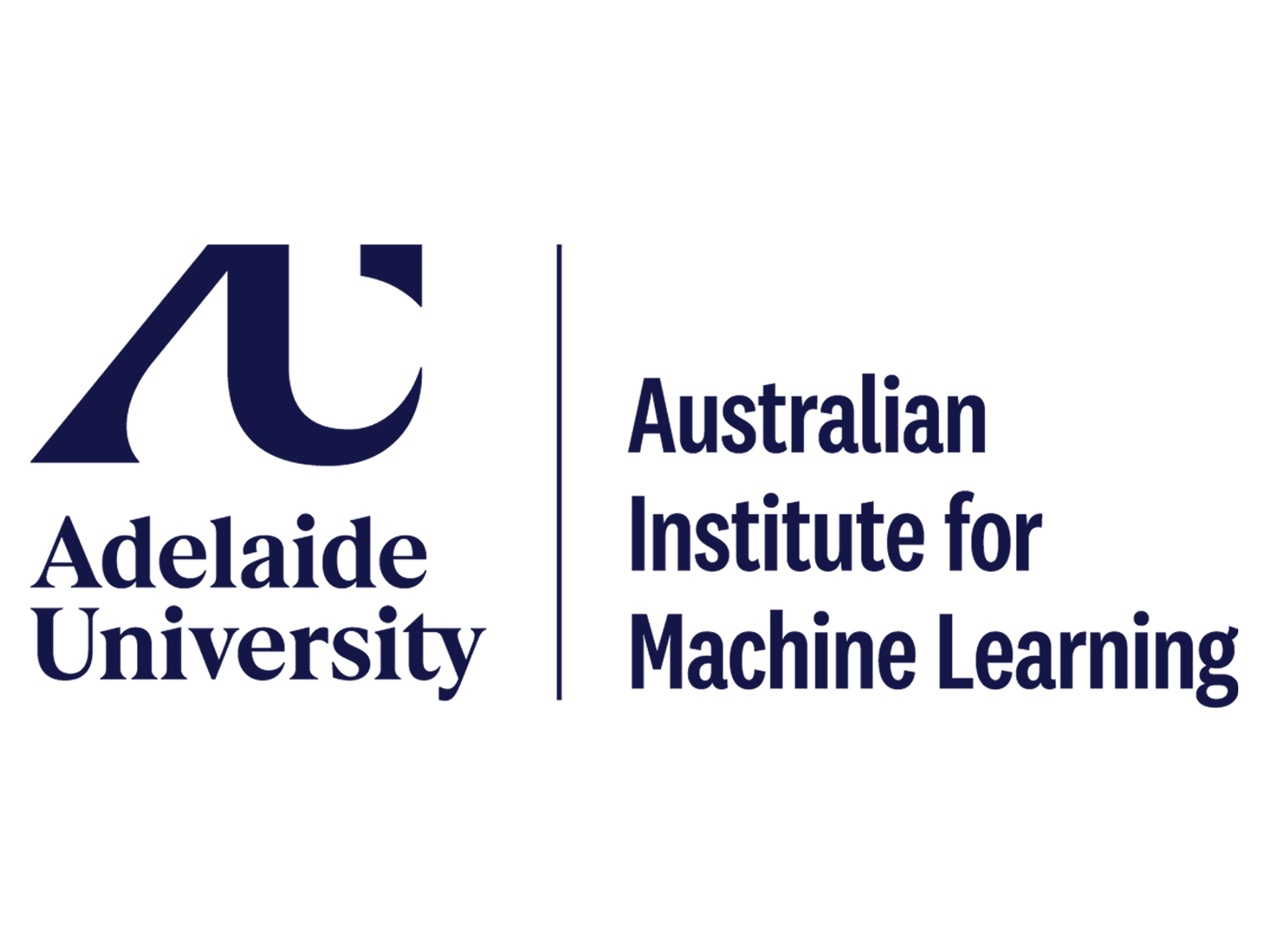}}}\hfill
\makebox[0.31\linewidth][c]{\raisebox{-0.5\height}{\includegraphics[height=0.95cm,viewport=104 520 1816 921,clip]{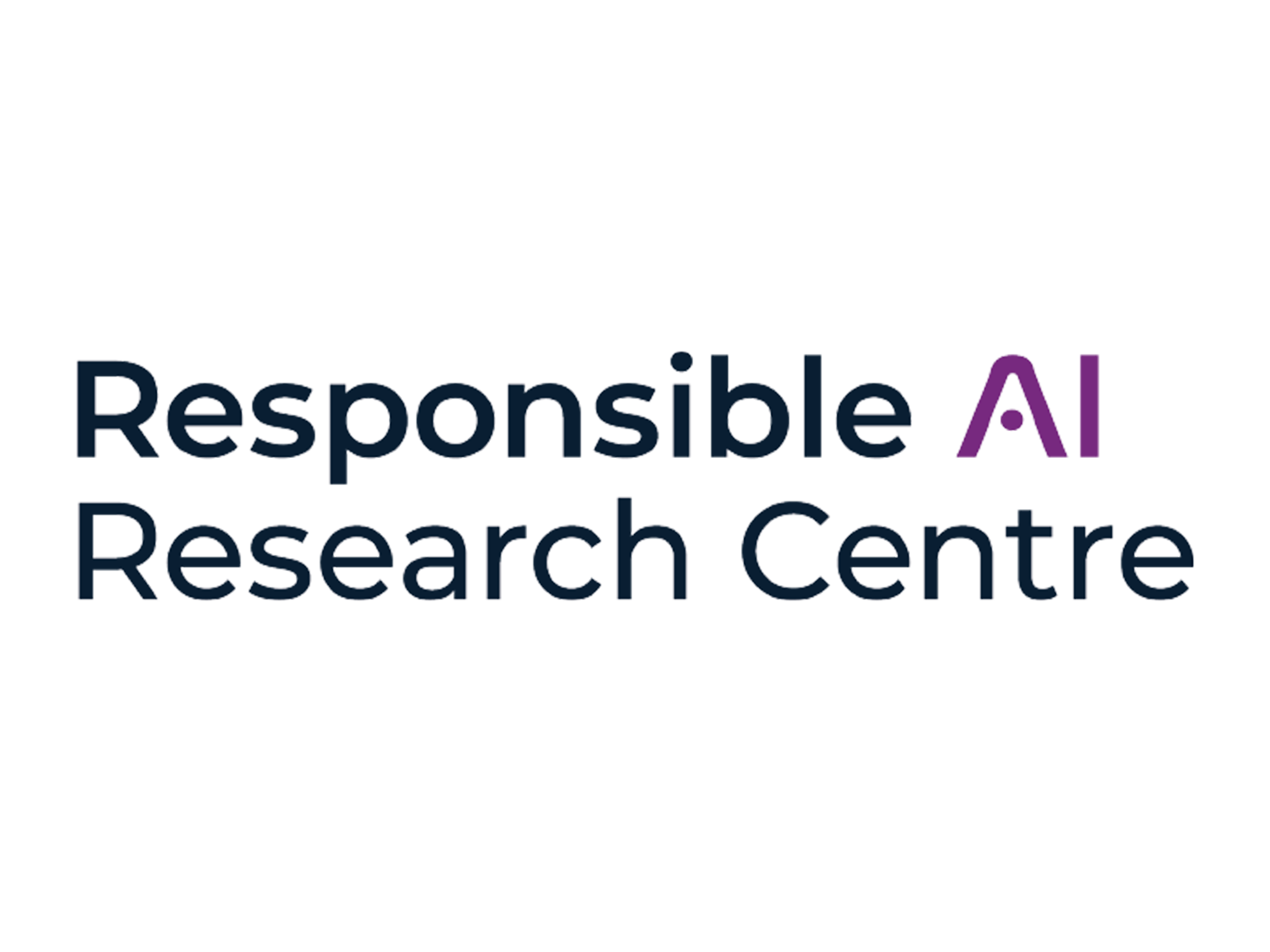}}}\hfill
\makebox[0.31\linewidth][c]{\raisebox{-0.5\height}{\includegraphics[height=1.55cm]{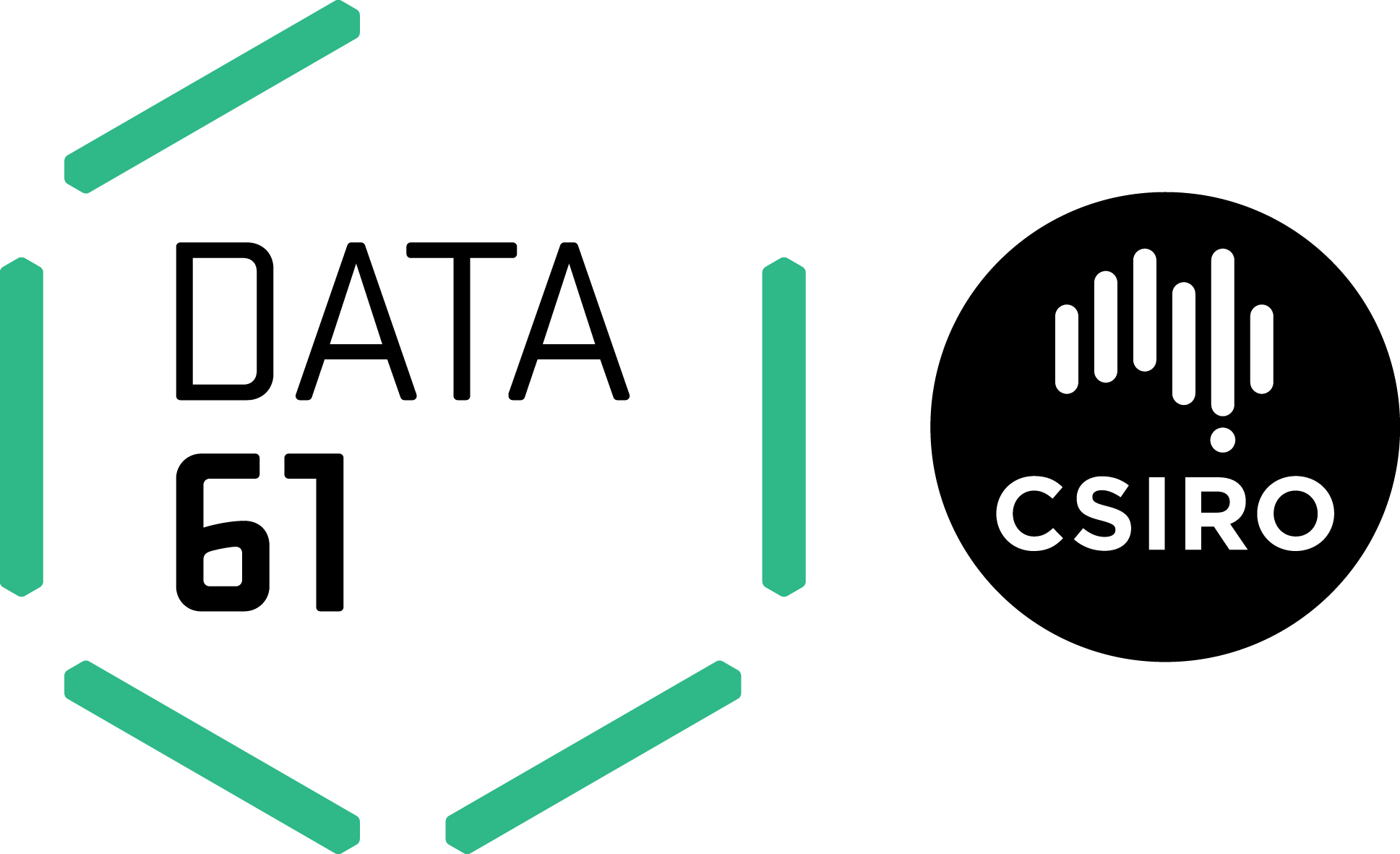}}}%
}\par\vspace{10pt}
\maketitle

\section{Introduction}
\label{sec:intro}

\begin{figure*}[t]
  \centering
  \includegraphics[width=\textwidth]{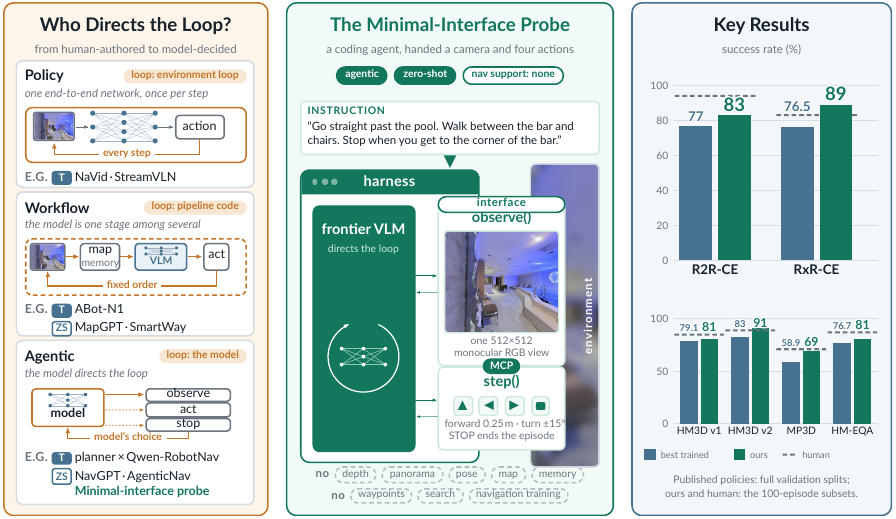}
  \caption{\textbf{Who directs the interaction loop.} Left: policy, workflow, and agentic control (T: trained; ZS: zero-shot). Middle: our minimal interface. Right: our best agent, published trained policies, and human reference (\cref{tab:board6}).}
  \label{fig:taxonomy-minimal}
\end{figure*}

Consider a household robot expected to operate autonomously day after day. It
must find the kitchen, pick up a cup, and recover from mistakes without
human rescue. Navigation and manipulation require different low-level
controllers, but share the high-level challenge of reasoning over extended
interaction. We ultimately seek an \emph{Autonomous Embodied Agent} that
can operate over extended interaction without repeated human rescue while
managing its state, errors, and computational resources.

Embodied navigation systems differ in who directs the interaction
loop (\cref{fig:taxonomy-minimal}, left). With a \emph{trained end-to-end
policy}, an external rollout loop repeatedly supplies one or more observation
frames and the task instruction to the model, executes the predicted action
or action chunk, and collects the next observation. NaVid and StreamVLN
learn navigation actions from video and language~\citep{zhang2024navid,wei2025streamvln},
much as Diffusion Policy and $\pi_0$ learn action policies for
manipulation~\citep{chi2023diffusionpolicy,black2024pi0}.
Industrial-scale models such as Qwen-RobotNav scale this policy-learning
approach through large-scale trajectory supervision and vision--language
co-training~\citep{zhang2026qwenrobotnav}.
The policy determines the actions; the rollout loop repeatedly queries the
policy and executes its output in a fixed observe--act cycle. Many zero-shot systems instead place
a frozen model inside a \emph{hand-designed workflow}. These workflows run
within the same rollout loop, but expand a single policy call into a sequence
of perception, memory, planning, and verification operations. MapGPT and DiscussNav build
such navigation-specific procedures around model
calls~\citep{chen2024mapgpt,long2024discussnav}. These operations are usually
organized into a predefined graph of nodes and
connections~\citep{zhou2026agentcanvas}. This gives the model more room to
reason, while the workflow still prescribes how that reasoning proceeds,
limiting its ability to handle situations outside the anticipated paths.
Some dual-system models, such as ABot-N1 and InternVLA-N1, also follow
predefined workflows, coupling a slow reasoning model to a trained action
expert through a fixed handoff~\citep{gong2026abotn1,wei2025dualvln}.
With \emph{agentic} control, the model chooses how the interaction proceeds.
NavGPT~\citep{zhou2024navgpt} provides an early example, implementing
ReAct~\citep{yao2023react} through a LangChain agent loop. More recently,
AgenticNav~\citep{li2026agenticnav} adopts a tool-calling harness.
From the goal and accumulated
feedback, it decides whether to observe again, act, check a result, revise
its plan, or stop. The harness executes calls and returns their results,
while the sequence of task operations is determined by the model at runtime.

For trained end-to-end policies and hand-designed workflows, much of the
interaction structure remains externally specified. This organization
deserves reconsideration as frontier vision--language models improve in
multi-turn, multimodal reasoning and spatial understanding. These
capabilities raise the possibility that the model can decide not only what
action to take, but also when to gather more information, check progress,
or revise its approach. \emph{How much of embodied interaction can the model
direct through a minimal interface?}

\looseness=-1 We study this question through embodied navigation and
question answering. We give thirteen frontier multimodal models, as
provided by their vendors, a minimal
interface---a monocular RGB view and four discrete actions---and let each
model decide when to observe, act, recover, and stop.
We call this organization \emph{agentic
embodied control}. We evaluate this setup across six benchmarks spanning
instruction-following navigation, object-goal navigation, and embodied
question answering.

\looseness=-1 This minimal setup yields strong performance across six
benchmarks. On the evaluated R2R-CE subset, the best agents achieve
82--83\% success, outperforming earlier zero-shot workflows on the same
episodes. Model choice matters more than harness complexity: the lightweight
mini-SWE-agent harness performs comparably to vendor harnesses, while earlier
workflows using the same top-performing models achieve substantially lower
success rates. Beyond this minimal interface, tools
help when the model can choose how to use them.
Offering a trained waypoint predictor and SLAM tools alongside primitive
actions improves the strongest agent, whereas imposing the predictor as
its action interface does not. Yet navigation failures, latency, context
growth, and difficulties in physical deployment still limit sustained
autonomy. These results show both the scope and the current limits of
model-directed embodied control.

\section{Related Work}
\label{sec:related}

\subsection{Learning and Scaffolding}
Recent navigation models differ in how they adapt pretrained visual and language
capabilities to control. NavFoM~\citep{zhang2025navfom} jointly trains on navigation
and image/video question answering across tasks and robot embodiments, using a
trajectory head for navigation. OmniNav~\citep{xue2025omninav} first trains discrete
actions and semantic tasks, then adds a flow-matching head for continuous waypoints
while retaining general-purpose training data. Robostral
Navigate~\citep{mistral2026robostral} combines large-scale simulated trajectory
supervision with online reinforcement learning to improve recovery and exploration.

Other systems provide spatial support at inference time. For zero-shot ObjectNav,
VLFM~\citep{yokoyama2024vlfm} uses a language-grounded value map to rank exploration
frontiers. GTA~\citep{li2026one} builds an explicit metric world representation and
presents rendered spatial views to an MLLM for waypoint selection.
P2DNav~\citep{sheng2026p2dnav} removes the learned waypoint predictor, but structures
decisions into panoramic direction selection and downview target grounding, supported
by dialogue memory and reflective reorientation. 

Embodied question answering further requires deciding where to seek evidence and when
to answer. Explore until Confident~\citep{ren2024explore} combines semantic mapping
with calibrated stopping, while GraphEQA~\citep{saxena2024grapheqa} uses online 3D
scene graphs and relevant images as memory for hierarchical planning.
FAST-EQA~\citep{zhang2026fasteqa} focuses exploration on question-relevant regions
and maintains a bounded scene memory. 

\begin{figure*}[t]
  \centering
  \includegraphics[width=\textwidth]{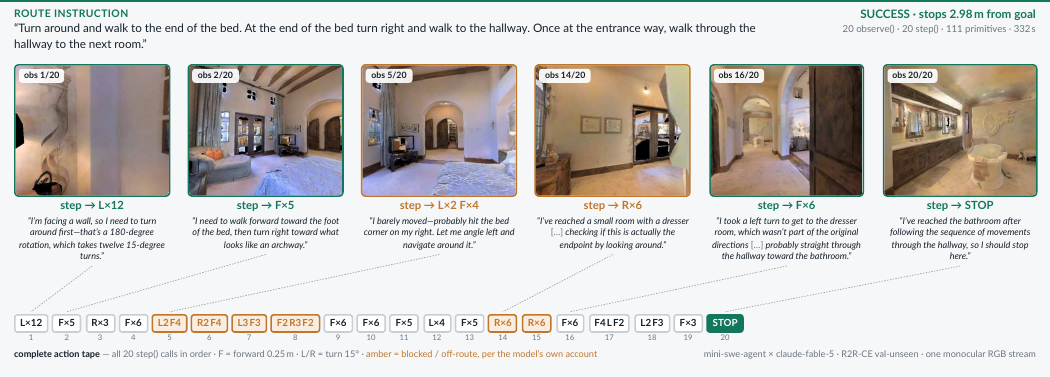}
  \caption{\textbf{A logged R2R-CE episode.} \texttt{fable-5} with mini-swe-agent: six of 20 observations with verbatim reasoning and calls; bottom: all 111 primitives. Amber marks model-reported blockage or route deviations.}
  \label{fig:react-episode}
\end{figure*}

\subsection{Model-Directed Embodied Interaction}
Embodied agents use tools for action and memory. In Minecraft,
Voyager~\citep{wang2023voyager} builds a reusable library of executable
skills and refines programs through environment feedback.
AgenticNav~\citep{li2026agenticnav} exposes pixel-target motion, depth queries, and
selective visual recall, allowing the model to request spatial evidence and revisit
observations. VIA~\citep{hu2026via} uses coding agents to manipulate
6-DoF waypoints through a visual interface.

Gemini Robotics 1.5~\citep{gemini2025robotics15} combines VLM orchestration
with a VLA executor, but its report does not establish a ReAct loop.
VLAs-as-Tools~\citep{lei2026vlastools} lets a VLM select trained controllers
and replan from execution feedback. We study how model capability, harness
choice, and navigation tools affect embodied interaction through a minimal
visual--action interface.

\section{Agentic Embodied Control through a Minimal Interface}
\label{sec:agent}

We connect a general-purpose multimodal agent to the environment through
observation and action tools. The model chooses the next tool call from the
interaction history, while the harness executes it and returns the result
(\cref{fig:react-episode}). We use a simple synchronous loop, with a fresh
agent session for each episode; asynchronous, parallel, and event-triggered
arrangements are also possible. We first describe the minimal interface,
then introduce the waypoint predictor and SLAM tools.

\subsection{A Minimal Perception--Action Interface}
\label{sec:interface}

For Room-to-Room (R2R-CE) and Room-across-Room (RxR-CE), the
embodied interface only contains two tools.
Both are served over the Model Context Protocol (MCP)~\citep{anthropic2024mcp}.
\texttt{observe()} returns a single
$512{\times}512$ front-facing RGB frame without advancing the environment.
\texttt{step(actions)} executes a bounded sequence of
primitive actions:
\texttt{FORWARD}, \texttt{LEFT}, \texttt{RIGHT}, and
\texttt{STOP}, which ends the episode. The tool reports the number of executed
actions and the remaining budget, but no image or collision signal, so the agent
must call \texttt{observe()} to see an action's effect.
These tools, with a briefing that states the instruction, the action
table, and the budget (\cref{app:interfaces}), form the entire task-specific
interface. The agent receives no pose, depth, panorama, map, or waypoint candidates.
ObjectNav and EQA additionally provide two camera actions:
look up and look down. Task-specific settings and the original tool
descriptions are provided in \cref{app:interfaces}.

\subsection{Waypoint and SLAM Tools}
\label{sec:slam-instruments}
\label{sec:interfaces}
We extend the minimal interface with a trained waypoint predictor and SLAM tools to examine how additional navigation support affects agent performance. These tools rely on sensor observations; no ground-truth poses or maps are used during navigation.

 \textbf{Waypoint interfaces.} We use the trained waypoint predictor adopted by SmartWay~\citep{shi2025smartway,hong2022bridging} to propose candidate waypoints from panoramic observations. In the \textbf{Waypoint} setting, the agent selects among these candidates to move. In \textbf{Hybrid}, waypoint tools are available alongside primitive actions, allowing the agent to choose either.

\textbf{SLAM tools.} An ORB-SLAM3~\citep{campos2021orbslam3} backend processes RGB-D observations to estimate pose and build a map. Tools expose these estimates, trajectories, and path previews, alongside panoramic observations and location markers. Adding this toolset to Minimal or Hybrid yields \textbf{SLAM} and \textbf{Hybrid + SLAM}, respectively. Full tool descriptions and budgets are provided in \cref{app:interfaces}.

\section{Experimental Evaluation}
\label{sec:capability}
\label{sec:setup}
\label{sec:protocol}

Our main experiments evaluate how well general-purpose agents can act
through a minimal embodied interface. We test three task families,
vision-and-language navigation (VLN), object-goal navigation (ObjectNav), and
embodied question answering (EQA), across six benchmarks:
R2R-CE~\citep{krantz2020beyond} and RxR-CE~\citep{ku2020rxr} for VLN;
HM3D-ObjectNav v1, HM3D-ObjectNav v2, and
MP3D-ObjectNav~\citep{batra2020objectnav} for ObjectNav; and
HM-EQA~\citep{ren2024explore} for EQA.

\paragraph{Splits and protocol.}
 We evaluate each benchmark on one fixed 100-episode subset shared by all models. For R2R-CE, we use the commonly used \texttt{rand100} val-unseen sample introduced by Open-Nav~\citep{qiao2025opennav}. The other five benchmarks have no agreed zero-shot subset. To provide a standardized evaluation set for each, we construct \texttt{mip100} from its official validation split, stratified by scene and selected by a hash of the episode identity. This deterministic procedure allows anyone with the official files to reconstruct the same 100 episodes (\cref{app:benchmark}). These fixed subsets make comparisons reproducible while keeping API-based evaluation tractable.

 We first compare fourteen models using the minimal interface (\cref{sec:main}), then examine the effects of waypoint and SLAM tools (\cref{sec:ablation-waypoint}) and harness choice (\cref{sec:ablation-harness}). We further analyze inference cost (\cref{sec:long-horizon}) and failure modes (\cref{sec:failure-analysis}), followed by qualitative deployment on a physical robot (\cref{sec:real-robot}). Evaluation configurations are frozen before each run (\cref{app:frozen}). The main comparison uses each vendor's default reasoning level (\cref{app:thinking}); task-specific tools and budgets are detailed in \cref{app:interfaces}.

\subsection{Minimal Interface: Main Results}
\label{sec:main}

How much embodied interaction can a frontier model direct through a minimal interface? We test whether the model can decide when to observe, act, and stop using only a short task briefing and observation--action tools, without an external map, memory module, or navigation-specific planner (\cref{sec:interface}). \Cref{tab:board6} reports results for fourteen frontier multimodal models across six benchmarks. OpenAI and Claude models use Codex and Claude Agent SDK, respectively; \texttt{qwen3.8-max} uses mini-swe-agent. Each human result comes from one tester using the same episodes and interface. Published results provide context, but use each system's own evaluation setting and are not matched comparisons.

\begin{table*}[t]
  \centering
  \caption{\textbf{Six-benchmark results.} Ours: zero-shot, minimal interface, 100 episodes per benchmark; published results as reported. ObjectNav SR uses $1$\,m; SR$_{\mathrm{vis}}$ is visibility-adjusted; Acc: EQA accuracy. Bold/underline: best ours/published; Human unranked. $^{*}$: three-run mean $\pm$ s.d.; otherwise one run; --: unavailable. Protocol and row notes: \cref{app:benchmark}.}
  \label{tab:board6}
  \footnotesize
  \setlength{\tabcolsep}{4pt}
  \renewcommand{\arraystretch}{0.88}  %
  \resizebox{\textwidth}{!}{%
  \begin{tabular}{@{}l l r r p{4pt} c c p{4pt} c c p{4pt} c c p{4pt} r@{}}
    \toprule
    & & & & & \multicolumn{8}{c}{ObjectNav} & & \\
    \cmidrule(lr){6-13}
    & & R2R-CE & RxR-CE & & \multicolumn{2}{c}{HM3D v1} & & \multicolumn{2}{c}{HM3D v2} & & \multicolumn{2}{c}{MP3D} & & HM-EQA \\
    \cmidrule(lr){3-3}\cmidrule(lr){4-4}\cmidrule(lr){6-7}\cmidrule(lr){9-10}\cmidrule(lr){12-13}\cmidrule(l){15-15}
    System & Source & SR & SR & & SR & SR$_{\mathrm{vis}}$ & & SR & SR$_{\mathrm{vis}}$ & & SR & SR$_{\mathrm{vis}}$ & & Acc \\
    \midrule
    \multicolumn{15}{@{}l}{\emph{Published systems, full validation splits}} \\
    NaVid~\citep{zhang2024navid}$^{h}$ & trained & 37.4 & 23.8 &  & 32.5 & -- &  & -- & -- &  & -- & -- &  & -- \\
    \rowcolor{gray!9}
    Uni-NaVid~\citep{zhang2024uninavid} & trained & 47.0 & 48.7 &  & 73.7 & -- &  & -- & -- &  & -- & -- &  & -- \\
    NaVILA~\citep{cheng2024navila} & trained & 54.0 & 49.3 &  & -- & -- &  & -- & -- &  & -- & -- &  & -- \\
    \rowcolor{gray!9}
    Vesta~\citep{bjorck2026vesta} & trained & 55.5 & -- &  & -- & -- &  & -- & -- &  & -- & -- &  & -- \\
    StreamVLN~\citep{wei2025streamvln} & trained & 56.4 & 54.4 &  & -- & -- &  & -- & -- &  & -- & -- &  & -- \\
    \rowcolor{gray!9}
    Hy-Embodied-VLM (A3B)~\citep{wang2026hyembodied}$^{h}$ & trained & 57.9 & -- &  & -- & -- &  & -- & -- &  & 38.3 & -- &  & -- \\
    RynnBrain-Nav (8B)~\citep{dang2026rynnbrain} & trained & 58.6 & 56.1 &  & -- & -- &  & -- & -- &  & -- & -- &  & -- \\
    \rowcolor{gray!9}
    InternVLA-N1 / DualVLN~\citep{wei2025dualvln} & trained & 64.3 & 61.4 &  & -- & -- &  & -- & -- &  & -- & -- &  & -- \\
    OmniNav~\citep{xue2025omninav} & trained & 69.5 & 73.6 &  & -- & -- &  & -- & -- &  & -- & -- &  & -- \\
    \rowcolor{gray!9}
    ABot-N1~\citep{gong2026abotn1} & trained & 70.9 & 73.9 &  & -- & -- &  & -- & -- &  & -- & -- &  & -- \\
    Qwen-RobotNav (w/o its planner)~\citep{zhang2026qwenrobotnav}$^{b}$ & trained & 72.1 & \smash{\underline{76.5}} &  & -- & -- &  & 71.2 & -- &  & 48.8 & -- &  & 76.7 \\
    \rowcolor{gray!9}
    Robostral Navigate~\citep{mistral2026robostral}$^{a}$ & trained & \smash{\underline{77.4}} & 75.1 &  & -- & -- &  & -- & -- &  & -- & -- &  & -- \\
    FOM-Nav~\citep{chabal2025fomnav} & trained & -- & -- &  & 57.3 & -- &  & 75.8 & -- &  & 35.0 & -- &  & -- \\
    \rowcolor{gray!9}
    VLingNav~\citep{wang2026vlingnav} & trained & -- & -- &  & \smash{\underline{79.1}} & -- &  & \smash{\underline{83.0}} & -- &  & \smash{\underline{58.9}} & -- &  & -- \\
    \midrule
    DSCD-Nav~\citep{an2026dscdnav}$^{c}$ & zero-shot & -- & -- &  & 75.6 & -- &  & 73.0 & -- &  & 47.8 & -- &  & -- \\
    \rowcolor{gray!9}
    ApexNav~\citep{zhang2025apexnav} & zero-shot & -- & -- &  & 59.6 & -- &  & 76.2 & -- &  & 39.2 & -- &  & -- \\
    STRIVE~\citep{zhu2025strive}$^{d}$ & zero-shot & -- & -- &  & 62.9 & -- &  & 79.6 & -- &  & 52.3 & -- &  & -- \\
    \rowcolor{gray!9}
    PIGEON~\citep{peng2025pigeon} & zero-shot & -- & -- &  & 67.1 & -- &  & 79.6 & -- &  & 52.6 & -- &  & -- \\
    Explore-EQA~\citep{ren2024explore}$^{g}$ & zero-shot & -- & -- &  & -- & -- &  & -- & -- &  & -- & -- &  & 58.4 \\
    \rowcolor{gray!9}
    FAST-EQA~\citep{zhang2026fasteqa} & zero-shot & -- & -- &  & -- & -- &  & -- & -- &  & -- & -- &  & 69.2 \\
    ARNA~\citep{lange2025arna}$^{f}$ & zero-shot & -- & -- &  & -- & -- &  & -- & -- &  & -- & -- &  & \smash{\underline{77}} \\
    \midrule
    \multicolumn{15}{@{}l}{\emph{Published zero-shot agents, 100-episode subsets}} \\
    AgenticNav (GPT-5.5)~\citep{li2026agenticnav}$^{e}$ & zero-shot & 55 & -- &  & -- & -- &  & -- & -- &  & -- & -- &  & -- \\
    \rowcolor{gray!9}
    SmartWay (opus-5.5)~\citep{shi2025smartway}$^{e}$ & zero-shot & 65 & -- &  & -- & -- &  & -- & -- &  & -- & -- &  & -- \\
    O2C-Nav (opus-5.5)~\citep{pan2026o2cnav}$^{e}$ & zero-shot & 69 & -- &  & -- & -- &  & -- & -- &  & -- & -- &  & -- \\
    \midrule
    \emph{Human} & -- & 94 & 83 &  & 85 & 97 &  & 89 & 93 &  & 71 & 85 &  & 87 \\
    \midrule
    \multicolumn{15}{@{}l}{\emph{Minimal interface (ours), zero-shot, 100-episode subsets}} \\
    \texttt{gpt-5.6-luna} & Codex & $^{*}$38.0 $\pm$ 2.0 & 29 &  & 53 & 69 &  & 68 & 75 &  & 41 & 66 &  & 67.0 \\
    \rowcolor{gray!9}
    \texttt{gpt-6-luna} & Codex & $^{*}$40.3 $\pm$ 1.5 & 42 &  & 62 & 73 &  & 69 & 74 &  & 46 & 69 &  & 58.0 \\
    \texttt{gpt-5.5} & Codex & $^{*}$46.0 $\pm$ 3.0 & 42 &  & 61 & 78 &  & 65 & 83 &  & 46 & 74 &  & 60.0 \\
    \rowcolor{gray!9}
    \texttt{haiku-5.5} & SDK & $^{*}$49.0 $\pm$ 1.7 & 37 &  & 60 & 77 &  & 68 & 72 &  & 53 & 74 &  & 64.0 \\
    \texttt{sonnet-5} & SDK & $^{*}$51.3 $\pm$ 1.2 & 45 &  & 66 & 82 &  & 77 & 85 &  & 52 & 74 &  & 64.0 \\
    \rowcolor{gray!9}
    \texttt{qwen3.8-max} & mini & $^{*}$56.3 $\pm$ 3.1 & 43 &  & 61 & 81 &  & 67 & 75 &  & 51 & 73 &  & 52.0 \\
    \texttt{gpt-5.6} & Codex & $^{*}$56.7 $\pm$ 2.5 & 53 &  & 63 & 77 &  & 73 & 80 &  & 52 & 76 &  & 71.0 \\
    \rowcolor{gray!9}
    \texttt{fable-5} & SDK & $^{*}$68.3 $\pm$ 1.5 & 62 &  & 69 & 84 &  & 84 & 89 &  & 60 & 83 &  & 69.0 \\
    \texttt{gpt-6-sol} & Codex & $^{*}$69.7 $\pm$ 2.9 & 62 &  & 75 & 87 &  & 85 & 90 &  & 55 & 81 &  & 79.0 \\
    \rowcolor{gray!9}
    \texttt{opus-5} & SDK & $^{*}$70.7 $\pm$ 3.5 & 67 &  & 72 & 86 &  & 86 & 90 &  & 63 & 83 &  & 77.0 \\
    \texttt{gpt-6.1-sol} & Codex & $^{*}$75.0 $\pm$ 1.0 & 83 &  & 74 & 90 &  & 89 & \textbf{\smash{96}} &  & 62 & 86 &  & 79.0 \\
    \rowcolor{gray!9}
    \texttt{gpt-6-astra} & Codex & $^{*}$75.7 $\pm$ 4.7 & \textbf{\smash{89}} &  & \textbf{\smash{81}} & \textbf{\smash{94}} &  & \textbf{\smash{91}} & 95 &  & 66 & \textbf{\smash{89}} &  & 80.0 \\
    \texttt{opus-5.5} & SDK & $^{*}$82.3 $\pm$ 3.5 & 74 &  & 80 & 92 &  & 88 & 92 &  & \textbf{\smash{69}} & 88 &  & \textbf{\smash{81.0}} \\
    \rowcolor{gray!9}
    \texttt{fable-5.1} & SDK & \textbf{\smash{$^{*}$83.0 $\pm$ 1.7}} & 79 &  & 78 & 93 &  & 87 & 91 &  & 63 & 85 &  & 79.0 \\
    \bottomrule
  \end{tabular}}
\end{table*}

 \textbf{A minimal interface supports strong performance across tasks.} The leading models---\texttt{opus-5.5}, \texttt{fable-5.1}, \texttt{gpt-6-astra}, and \texttt{gpt-6.1-sol}---set the best scores across the six benchmarks, exceeding the listed published results on each. R2R-CE success ranges from 38.0 to 83.0 across models, showing that performance varies substantially even with the same task interface. The best models reach 82--83 SR against our human tester's 94 on R2R-CE; on RxR-CE, the best score is 89 against the tester's 83. HM-EQA accuracy reaches 81. These results show that model-directed observation and action can support both navigation and question answering with little task-specific scaffolding.

\textbf{The remaining failures reflect both agent limitations and evaluation criteria.} The R2R-CE case study (\cref{app:case-study}) identifies episodes in which the agent follows an incorrect interpretation of the instruction; some revisit earlier locations but still fail to reach the goal. On the three ObjectNav benchmarks, however, some apparent failures arise from incomplete annotations or the success criterion. A scene may contain an unannotated target instance, while success requires stopping near an annotated goal view point. Reaching an unlabeled instance, or viewing a labeled one from another side, can therefore count as a failure. For ObjectNav, SR uses a $1$\,m threshold around annotated goal viewpoints; SR$_{\mathrm{vis}}$ additionally credits stops with the target visible in the final observations (\cref{app:objnav-screen}). This distinction helps separate failures to find the target from failures to satisfy the annotated-viewpoint criterion. \Cref{sec:failure-analysis} examines these cases in detail.

We cannot verify whether these models were post-trained on the evaluated benchmarks, and our evaluation is limited to fixed 100-episode subsets. Nevertheless, their performance and observed trajectories are encouraging, suggesting that a minimal observation--action interface can already support capable embodied behavior. We next examine how additional tools and harness design affect this performance.

\subsection{Hybrid Interface with SLAM}
\label{sec:ablation-waypoint}
\label{sec:vlnverse}
\label{sec:hybrid}
\begin{table}[t]
  \centering
  \caption{\textbf{Waypoint-only control on R2R-CE.} Matched models, harnesses, and episodes. $^{*}$: replication mean; $^{\dagger}$: medium effort; bold: $|\Delta|\ge4$ SR.}
  \label{tab:waypoint}
  \label{tab:interface}
  \label{tab:ablation-waypoint}
  \small
  \setlength{\tabcolsep}{8pt}
  \renewcommand{\arraystretch}{1.0}
  {%
  \begin{tabular}{@{}ll rrr rr@{}}
    \toprule
    & & \multicolumn{3}{c}{SR$\uparrow$} & \multicolumn{2}{c}{SPL$\uparrow$} \\
    \cmidrule(lr){3-5}\cmidrule(l){6-7}
    Benchmark & Model & Primitive & Waypoint & $\Delta$ & Primitive & Waypoint \\
    \midrule
    R2R-CE & \texttt{qwen3.5-4b} & 5 & 43 & \textbf{+38} & 4.58 & 34.18 \\
    R2R-CE & \texttt{qwen3.5-9b} & 7 & 44 & \textbf{+37} & 5.36 & 32.81 \\
    R2R-CE & \texttt{qwen3.5-plus} & $^{*}$41.3 & 53 & \textbf{+11.7} & 31.81 & 44.05 \\
    \midrule
    R2R-CE & \texttt{gpt-5.5} & $^{*}$46.0 & 67 & \textbf{+21} & 40.13 & 58.85 \\
    R2R-CE & \texttt{gpt-5.6} & $^{*}$56.7 & $^{\dagger}$73 & \textbf{+16.3} & 41.78 & 63.98 \\
    \midrule
    R2R-CE & \texttt{haiku-5.5} & $^{*}$49.0 & 59 & \textbf{+10} & 39.88 & 49.28 \\
    R2R-CE & \texttt{sonnet-5} & $^{*}$51.3 & 60 & \textbf{+8.7} & 37.84 & 45.83 \\
    R2R-CE & \texttt{opus-4.8} & $^{*}$55.7 & 65 & \textbf{+9.3} & 47.31 & 54.23 \\
    R2R-CE & \texttt{fable-5} & $^{*}$68.3 & 69 & +0.7 & 58.02 & 59.15 \\
    R2R-CE & \texttt{opus-5} & $^{*}$70.7 & 70 & $-$0.7 & 55.21 & 58.27 \\
    R2R-CE & \texttt{opus-5.5} & $^{*}$82.3 & $^{*}$83.7 & +1.4 & 71.28 & 71.05 \\
    R2R-CE & \texttt{fable-5.1} & $^{*}$83.0 & 81 & $-$2 & 71.25 & 69.93 \\
    \bottomrule
  \end{tabular}}
\end{table}

\begin{table}[t]
  \centering
  \caption{\textbf{Interfaces on R2R-CE.} Turns, Wall, USD: per-episode calls, time, and vendor-priced cost relative to each model's Minimal baseline. Bold: best SR; $^{*}$: three-run mean $\pm$ s.d.; otherwise one run; --: not run.}
  \label{tab:interfaces-r2r}
  \small
  \setlength{\tabcolsep}{8pt}
  \renewcommand{\arraystretch}{1.0}
  {%
  \begin{tabular}{@{}ll rrrr rrr@{}}
    \toprule
    & & & & & & \multicolumn{3}{c}{$\times$ Minimal} \\
    \cmidrule(l){7-9}
    Model & Interface & SR$\uparrow$ & SPL$\uparrow$ & NE$\downarrow$ & OSR$\uparrow$ & Turns & Wall & USD \\
    \midrule
    \texttt{haiku-5.5} & Minimal & $^{*}$49.0 $\pm$ 1.7 & 39.88 & 5.63 & 58.3 & 1 & 1 & 1 \\
     & Waypoint & \textbf{59} & 49.28 & 5.11 & 67 & 0.43 & 0.42 & 0.55 \\
     & Hybrid & 51 & 42.31 & 5.24 & 59 & 0.58 & 0.58 & 0.78 \\
     & SLAM & 51 & 38.65 & 5.54 & 59 & 1.12 & 1.60 & 1.88 \\
     & Hybrid + SLAM & 56 & 45.14 & 5.29 & 64 & 0.67 & 0.93 & 1.05 \\
    \midrule
    \texttt{fable-5} & Minimal & $^{*}$68.3 $\pm$ 1.5 & 58.02 & 5.13 & 73.3 & 1 & 1 & 1 \\
     & Waypoint & 69 & 59.15 & 4.30 & 77 & 0.34 & 0.38 & 0.27 \\
     & Hybrid & 77 & 65.45 & 3.35 & 83 & 0.47 & 0.52 & 0.40 \\
     & Hybrid + SLAM & \textbf{78} & 66.13 & 3.27 & 81 & 0.53 & 0.78 & 0.60 \\
    \midrule
    \texttt{opus-5} & Minimal & $^{*}$70.7 $\pm$ 3.5 & 55.21 & 4.79 & 78.3 & 1 & 1 & 1 \\
     & Waypoint & 70 & 58.27 & 4.03 & 81 & 0.45 & 0.56 & 0.47 \\
     & Hybrid & 78 & 62.62 & 3.26 & 86 & 0.54 & 0.59 & 0.60 \\
     & SLAM & 76 & 59.82 & 3.54 & 82 & 0.70 & 0.81 & 0.68 \\
     & Hybrid + SLAM & \textbf{80} & 64.55 & 2.68 & 86 & 0.65 & 0.81 & 0.78 \\
    \midrule
    \texttt{opus-5.5} & Minimal & $^{*}$82.3 $\pm$ 3.5 & 71.28 & 3.03 & 84.7 & 1 & 1 & 1 \\
     & Waypoint & $^{*}$83.7 $\pm$ 1.2 & 71.05 & 2.28 & 88.0 & 0.57 & 0.53 & 0.81 \\
     & Hybrid & $^{*}$85.7 $\pm$ 1.5 & 75.32 & 2.65 & 90.0 & 0.59 & 0.62 & 0.88 \\
     & SLAM & $^{*}$82.7 $\pm$ 0.6 & 69.89 & 2.82 & 87.0 & 0.74 & 1.11 & 1.16 \\
     & Hybrid + SLAM & \textbf{$^{*}$88.0 $\pm$ 1.0} & 75.38 & 2.09 & 91.7 & 0.64 & 0.96 & 1.05 \\
    \midrule
    \texttt{fable-5.1} & Minimal & $^{*}$83.0 $\pm$ 1.7 & 71.25 & 2.70 & 86.7 & 1 & 1 & 1 \\
     & Waypoint & 81 & 69.93 & 2.74 & 83 & 0.48 & 0.42 & 0.68 \\
     & Hybrid & 83 & 72.09 & 2.78 & 85 & 0.58 & 0.49 & 0.82 \\
     & SLAM & 79 & 65.99 & 2.84 & 83 & 0.78 & 1.04 & 1.31 \\
     & Hybrid + SLAM & \textbf{86} & 72.95 & 2.64 & 88 & 0.64 & 0.76 & 1.06 \\
    \bottomrule
  \end{tabular}}
\end{table}

 \textbf{Waypoints boost weaker models; stronger models benefit more when waypoints are optional rather than imposed.} We first replace primitive movement with the waypoint predictor described in \cref{sec:slam-instruments}, so the agent moves by selecting its proposed candidates. This substantially helps models with lower baseline success: \texttt{qwen3.5-4b} rises from 5 to 43 SR and \texttt{qwen3.5-9b} from 7 to 44. For stronger models, the benefit largely disappears: \texttt{fable-5}, \texttt{opus-5}, \texttt{opus-5.5}, and \texttt{fable-5.1} change by at most two points in either direction (\cref{tab:waypoint}). However, keeping primitive actions alongside waypoints changes the outcome. In this Hybrid setting, the model chooses how to move rather than relying exclusively on the predictor. We examine this setting and the SLAM variants below primarily on \texttt{opus-5.5}, with three runs per setting. Single-run tool comparisons on \texttt{haiku-5.5}, \texttt{fable-5}, \texttt{opus-5}, and \texttt{fable-5.1} check whether the trends extend to other models, using their repeated-run Minimal baselines (\cref{tab:interfaces-r2r}). Hybrid raises \texttt{opus-5.5} from 82.3 to 85.7 SR while using 0.59$\times$ the model calls, 0.62$\times$ the wall time, and 0.88$\times$ the cost of Minimal. The supplementary comparisons show gains for \texttt{fable-5} (68.3 to 77) and \texttt{opus-5} (70.7 to 78); \texttt{fable-5.1} stays at 83. These four stronger models achieve higher SR with Hybrid than with waypoint-only control.

 \textbf{Hybrid + SLAM achieves the highest SR for stronger models.} Maps are central to many embodied navigation systems, where semantic and value maps guide exploration and goal selection. We therefore ask whether an estimated map also helps a model-directed agent. Our ORB-SLAM3 backend builds a map and estimates pose from RGB-D observations, with no ground-truth poses or maps used during navigation. Adding these tools to Minimal gives mixed results: \texttt{opus-5} improves from 70.7 to 76 SR, \texttt{opus-5.5} changes little (82.3 to 82.7), and \texttt{fable-5.1} falls from 83 to 79. Combined with Hybrid, however, SLAM increases SR across the five tested models. On \texttt{opus-5.5}, SR rises from 85.7 to 88.0 and navigation error falls from 2.65 to 2.09\,m. The supplementary runs reach 78, 80, and 86 SR on \texttt{fable-5}, \texttt{opus-5}, and \texttt{fable-5.1}, respectively, each above its Hybrid score and the highest SR for that model. On \texttt{haiku-5.5}, SLAM also improves Hybrid from 51 to 56 SR, but waypoint-only control still achieves higher SR at 59. These gains come with additional overhead: adding SLAM to Hybrid increases model calls, wall time, and cost for all five models. For \texttt{opus-5.5}, Hybrid + SLAM takes 0.96$\times$ the wall time and costs 1.05$\times$ as much as Minimal. Thus, Hybrid + SLAM achieves the highest SR for the stronger models tested here, while Hybrid offers a less costly alternative. The weaker \texttt{haiku-5.5} instead achieves its highest SR with waypoint-only control.

\subsection{Harness: Who Directs the Loop Matters}
\label{sec:ablations}
\label{sec:ablation-harness}
\label{sec:ablation-director}

 \textbf{Stronger models perform better in the agent loop.} Can a fixed workflow obtain the same performance from the same model? We test SmartWay~\citep{shi2025smartway} and O2C-Nav~\citep{pan2026o2cnav}, which combine a base model with a waypoint module through predefined task operations. Unlike the agent loop, these workflows determine when the model is queried and how its outputs are used (\cref{fig:taxonomy-minimal}). We run both on the same R2R-CE episodes with seven of the models in \cref{tab:board6}. The results depend on model capability (\cref{tab:harness-r2r}). For \texttt{gpt-5.5}, SmartWay and O2C-Nav reach 49 and 54 SR, respectively, against 46.0 with the native agent harness. For stronger models, the pattern reverses: \texttt{opus-5.5} falls from 82.3 to 65 and 69, and \texttt{fable-5.1} from 83.0 to 62 in both workflows. The same direction holds for \texttt{gpt-6-sol} and \texttt{gpt-6-astra}. Thus, a stronger base model does not automatically translate into stronger navigation within a fixed workflow; the model-directed agents make better use of the stronger models in these comparisons.

\begin{figure*}[!t]
 \centering
 \includegraphics[width=\textwidth]{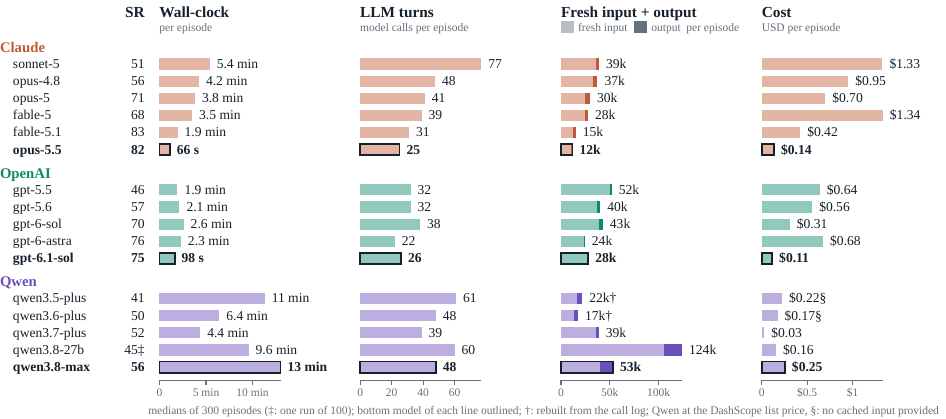}
 \caption{\textbf{Runtime and cost on R2R-CE.} Per-episode medians with the Minimal interface and default effort; SR beside each model (\cref{app:cost}).}
 \label{fig:cost}
\end{figure*}

 \textbf{A simple harness retains strong performance.} Could this advantage instead come from the vendor harnesses? We replace Claude Agent SDK and Codex with mini-swe-agent~\citep{minisweagent2025}, a lightweight coding-agent harness that repeatedly queries the model, executes its tool calls, and returns the results. We retain the same minimal embodied interface and evaluation episodes. Across twelve models, its SR differs from the native harness by roughly $-8$ to $+8$ points, with gains for some models and losses for others (\cref{tab:harness-r2r}). For the leading models, the differences are small: \texttt{fable-5.1} moves from 83.0 to 84, \texttt{opus-5.5} from 82.3 to 80, and \texttt{gpt-6-astra} from 75.7 to 79. Serving and context-handling settings differ across harnesses (\cref{app:harness}), so this is a comparison of complete systems rather than an isolated test of harness complexity. Nevertheless, the results show that strong embodied performance does not require the vendor harnesses tested here: a simple loop that lets the model direct interaction is sufficient.

\begin{table}[t]
  \centering
  \caption{\textbf{Harnesses and workflows on R2R-CE.} Native/mini-swe use the minimal interface; SmartWay/O2C-Nav use waypoints. $\Delta$: SR difference from Native; $^{*}$: three-run mean $\pm$ s.d.; $^{\ddagger}$: AgenticNav reports 44; --: not run.}
  \label{tab:harness-r2r}
  \footnotesize
  \setlength{\tabcolsep}{8pt}
  \renewcommand{\arraystretch}{1.0}
  {%
  \begin{tabular}{@{}l | r | rr | rr | rr@{}}
    \toprule
     &  & \multicolumn{2}{c|}{mini-swe} & \multicolumn{2}{c|}{SmartWay} & \multicolumn{2}{c}{O2C-Nav} \\
    \cmidrule(lr){3-4}\cmidrule(lr){5-6}\cmidrule(l){7-8}
    Model & Native & SR & $\Delta$ & SR & $\Delta$ & SR & $\Delta$ \\
    \midrule
    \texttt{gpt-6-luna} & $^{*}$40.3 $\pm$ 1.5 & 48 & +8 & 38 & $-$2 & 41 & +1 \\
    \texttt{gpt-5.5} & $^{*}$46.0 $\pm$ 3.0 & 52 & +6 & $^{\ddagger}$49 & +3 & 54 & +8 \\
    \texttt{gpt-5.6} & $^{*}$56.7 $\pm$ 2.5 & 60 & +3 & 45 & $-$12 & 58 & +1 \\
    \texttt{gpt-6-sol} & $^{*}$69.7 $\pm$ 2.9 & 62 & $-$8 & 52 & $-$18 & 61 & $-$9 \\
    \texttt{gpt-6-astra} & $^{*}$75.7 $\pm$ 4.7 & 79 & +3 & 59 & $-$17 & 72 & $-$4 \\
    \midrule
    \texttt{haiku-5.5} & $^{*}$49.0 $\pm$ 1.7 & 47 & $-$2 & -- & -- & -- & -- \\
    \texttt{sonnet-5} & $^{*}$51.3 $\pm$ 1.2 & 53 & +2 & -- & -- & -- & -- \\
    \texttt{opus-4.8} & $^{*}$55.7 $\pm$ 2.3 & 63 & +7 & -- & -- & -- & -- \\
    \texttt{fable-5} & $^{*}$68.3 $\pm$ 1.5 & 72 & +4 & -- & -- & -- & -- \\
    \texttt{opus-5} & $^{*}$70.7 $\pm$ 3.5 & 69 & $-$2 & -- & -- & -- & -- \\
    \texttt{opus-5.5} & $^{*}$82.3 $\pm$ 3.5 & 80 & $-$2 & 65 & $-$17 & 69 & $-$13 \\
    \texttt{fable-5.1} & $^{*}$83.0 $\pm$ 1.7 & 84 & +1 & 62 & $-$21 & 62 & $-$21 \\
    \bottomrule
  \end{tabular}}
\end{table}

\subsection{Runtime and Cost}
\label{sec:long-horizon}
\label{sec:runtime-limitation}

 \Cref{fig:cost} reports per-episode medians of model calls, wall time, fresh input plus output tokens, and vendor-priced cost on R2R-CE, including failed episodes. Full accounting is in \cref{app:cost}.

 Newer frontier models generally complete episodes faster. From \texttt{opus-5} to \texttt{opus-5.5}, median time falls from 3.8\,min to 66\,s, calls from 41 to 25, and cost from \$0.70 to \$0.14, while SR rises from 70.7 to 82.3. From \texttt{gpt-5.5} to \texttt{gpt-6.1-sol}, median time falls from 116 to 98\,s. Qwen improves from 3.5-plus to 3.7-plus, but \texttt{qwen3.8-max} reverses this trend. API instability caused many Qwen episodes to reach the 2400\,s timeout, affecting both runtime and success rates.

\subsection{Failure Analysis}
\label{sec:behavior}
\label{sec:failure-analysis}

\begin{figure}[!htb]
 \centering
 \includegraphics[width=0.62\linewidth]{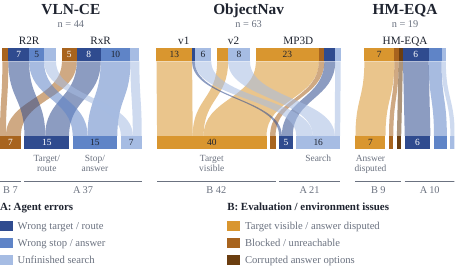}
 \caption{\textbf{Failure analysis of \texttt{opus-5.5}.} One run per benchmark; ObjectNav threshold: $1$\,m.  Flows connect benchmarks to diagnostic categories; widths count episodes within each panel.  Blue: agent errors; amber: evaluation or environment issues. Categories come from model-assisted review.}
 \label{fig:failure-flow}
\end{figure}

We examine archived logs and observations from one \texttt{opus-5.5} run per benchmark. Under the current criteria, including the $1$\,m ObjectNav threshold, there are 126 failures: 44 in VLN-CE, 63 in ObjectNav, and 19 in HM-EQA (\cref{fig:failure-flow}). The analysis is model-assisted without independent human verification (\cref{app:case-study,app:objnav-audit}).

\textbf{ObjectNav failures include visible targets.} Of the 63 failures, the audit labels 39 as target-visible and one as having a disputed category.  The remaining cases are five wrong targets, sixteen unfinished searches, and two unreachable goals. These cases motivate visibility-adjusted SR (\cref{app:objnav-screen}): an agent may see an unannotated target or view an annotated one from outside its accepted viewpoints. Visibility alone, however, does not establish a successful approach.

 \textbf{VLN failures often involve wrong targets, routes, or stops.}  Of the 44 VLN-CE failures, 15 involve a wrong target or route choice, 15 a wrong stopping location, seven an unfinished search, and seven blocked movement or other navigation obstacles. Some episodes revisit earlier locations but still fail to reach the goal.  These cases show difficulties in matching instructions to the scene and deciding where to stop, even when the agent revisits earlier locations.

 \textbf{EQA failures include wrong answers and disputed judgments.} Of the 19 rejected answers, six involve the wrong room or referent, three involve visual misinterpretation, one follows an unfinished search, one involves blocked movement, and one has corrupted answer options.  The seven cases labeled ``Answer disputed have answers that the audit finds defensible from the recorded views or questions it finds ambiguous. These cases need further review; they do not establish that the reference answers are wrong.  Across tasks, choosing the right target or route, stopping at the right place, and verifying answers remain recurring difficulties.

\subsection{Deployment on a Physical Robot}
\label{sec:real-robot}

We deploy the same minimal agent on a Unitree Go2 quadruped with the monocular observation and primitive-action interface used in simulation. We qualitatively examine 31 hand-designed episodes covering conditional navigation, counting, and multistage routes (\cref{app:real}). The agent resolves logical and perceptual conditions before acting, identifies a recipient by white shoes, and completes some return routes. Object pickup is assumed rather than executed.

Reliable control remains limited by body awareness, motion feedback, and persistent spatial state. Without body-state sensor information, the agent can turn too early and wedge the quadruped's rear, which extends behind the camera, in a doorway. Because \texttt{step} reports requested rather than realized motion, under-rotation can leave the agent following an incorrect heading. Cross-view counting and revisits to similar targets also expose difficulty maintaining spatial state through motion.

\FloatBarrier

\section{Conclusion}
\label{sec:conclusion}

We study how much embodied interaction frontier models can direct through a minimal observation--action interface. \textbf{(1)} Across six benchmarks, capable models achieve strong navigation and question-answering performance with little task-specific scaffolding. \textbf{(2)} Who directs the loop matters: with the same stronger models, model-directed agents outperform the predefined SmartWay and O2C-Nav workflows on R2R-CE. \textbf{(3)} A lightweight agent harness retains strong performance. \textbf{(4)} Stronger models benefit from optional rather than imposed waypoints, with SLAM further improving the Hybrid interface. These findings support agentic embodied control as a promising direction toward autonomous embodied agents. Reliable physical execution and persistent spatial understanding remain important next steps.

\section*{Limitations}
\label{sec:limitations}

We evaluate fixed 100-episode subsets of six benchmarks; scores may differ on the full sets. Published systems use different evaluation settings, and each human reference comes from one tester. We cannot verify whether proprietary models were post-trained on these benchmarks: zero-shot describes how we use them, not their training history.

Tool and harness comparisons mainly cover R2R-CE, with only one run for several supplementary settings. Small differences need further replication. Serving and context handling differ across harnesses, preventing us from isolating the effect of interaction structure. API instability also affects runtime and success, particularly for Qwen. Our model-assisted failure analysis has not been independently checked by humans, and target visibility alone does not confirm a successful approach.

The 31 physical trials are hand-designed and qualitative. Limited body-state and motion feedback affect execution, while instruction grounding, stopping, and maintaining spatial state remain difficult. These trials cover individual episodes; sustained autonomy and broader physical tasks remain untested.

\FloatBarrier
\bibliography{main}

\newpage
\clearpage
\appendix

\newcommand{\apxentry}[2]{%
  {\raggedright\hangindent=1.4em\hangafter=1\noindent\makebox[1.4em][l]{\textbf{\ref{#1}}}\textbf{#2}~\dotfill~\pageref{#1}\par}\smallskip}
{
\noindent\textbf{Contents of the Appendix}\par\smallskip
\apxentry{app:objnav-criterion}{ObjectNav False Negatives}
\apxentry{app:case-study}{Failure Analysis of One Opus 5.5 Run}
\apxentry{app:real}{Deployment on a Physical Robot}
\apxentry{app:aea}{From Episodic Embodied Agents to Autonomous Embodied Agents}
\apxentry{app:taxonomy}{A Taxonomy of Navigation Systems by Control Authority}
\apxentry{app:settings}{Experimental Settings}
}
\medskip

\section{ObjectNav False Negatives}
\label{app:objnav-criterion}

Some ObjectNav failures are false negatives: the agent finds and approaches a valid target, but its stop is rejected because the instance or the viewpoint it reaches is not annotated. We illustrate two cases and describe the supplementary visibility metric used in the main results.

\subsection{Examples}
\label{app:objnav-definition}

\begin{figure}[t]
  \centering
  \includegraphics[width=\columnwidth]{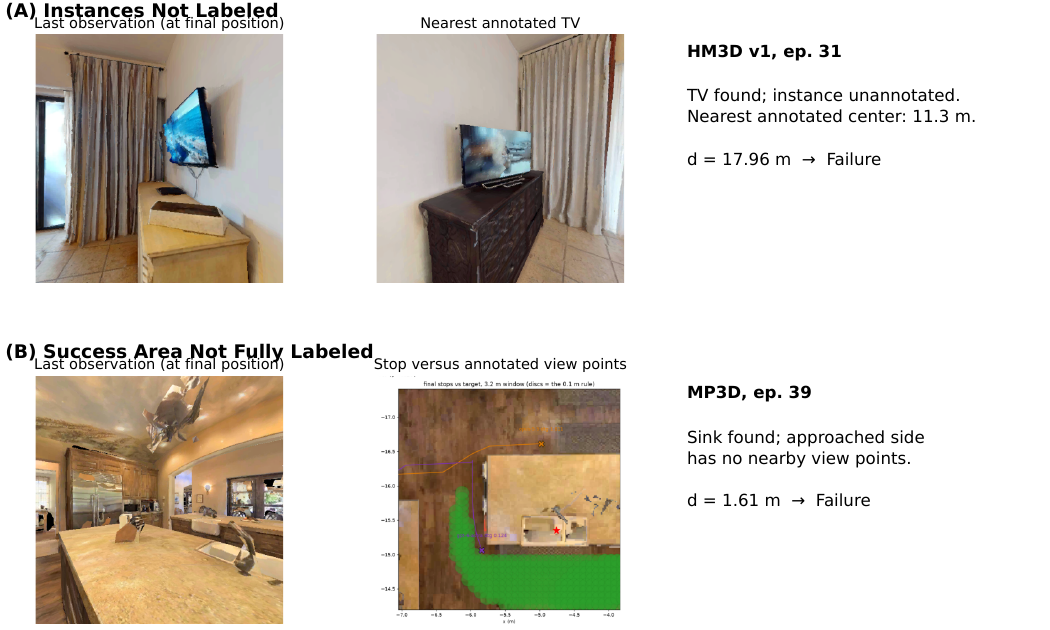}
  \caption{\textbf{Two false-negative patterns.} A: Instances Not Labeled. B: Success Area Not Fully Labeled. Both \texttt{opus-5.5} episodes are scored as failures. Left: observation after the final movement. Middle: an annotated-instance render (A) or top-down geometry (B). Right: rejection evidence. Green discs show $0.1$\,m Euclidean neighborhoods of view points; scoring uses geodesic distance. Red stars mark annotated centers; orange/purple crosses mark \texttt{opus-5.5}/\texttt{gpt-6-astra} stops.}
  \label{fig:objnav-issues}
\end{figure}

\Cref{fig:objnav-issues} shows two stops rejected despite a visible target. Here, $d_T$ is the geodesic distance to the nearest annotated goal viewpoint.

\textbf{Missing instance.} The agent stops in front of a television absent from the annotations; the nearest annotated television center is $11.3$\,m away, with $d_T=17.96$\,m.

\textbf{Missing viewpoints.} The agent stops beside a kitchen island with the annotated sink visible, but the accepted viewpoints lie on other sides, leaving $d_T=1.61$\,m.

\subsection{Metrics and Visual Screening}
\label{app:objnav-screen}

\textbf{Metrics.} Our ObjectNav SR counts \texttt{STOP} within $1$\,m of an annotated goal viewpoint, rather than the original evaluator\textquotesingle s $0.1$\,m threshold. SR$_{\mathrm{vis}}$ additionally counts stops at finite $d_T>1$\,m when the target category appears in any of the last three observations received before stopping. \emph{No}, \emph{unsure}, no-STOP, and unreachable cases remain failures. Published results are reported unchanged.

\textbf{Screening.} Claude Opus 5.5, run separately from the navigating agents, examines the goal category and up to three final observations and returns \emph{yes}, \emph{no}, or \emph{unsure}. It sees no maps, annotated instances, evaluator distances, or model identities; any visible instance of the requested category counts.  Across 48 runs, we screen 953 stops: 699 are judged visible, 129 not visible, and 125 uncertain.

\label{app:objnav-audit}
\textbf{Supporting audit.} For \texttt{opus-5.5} and \texttt{gpt-6-astra}, we also replayed 600 trajectories and checked 89 stops beyond $1$\,m using observations, replay geometry, and annotated-instance renders. This separate audit informs the failure analysis in \cref{sec:failure-analysis}. Neither audit has independent human verification. Visibility in observations preceding \texttt{STOP} does not establish successful approach; SR$_{\mathrm{vis}}$ is a supplementary measure of target finding.

\textbf{Human performance.} \Cref{tab:human-objnav} compares evaluator scores with volunteer self-assessments, using the same minimal interface and evaluation subsets as the model runs.

\begin{table}[!htb]
  \centering
  \caption{\textbf{Human ObjectNav performance (\%).} One volunteer without a research background per benchmark, $100$ episodes each. SR uses the listed distance thresholds; SR$_{\mathrm{vis}}$ adds target visibility. Self-assessed SR is volunteer-reported.}
  \label{tab:human-objnav}
  \footnotesize
  \setlength{\tabcolsep}{3pt}
  \begin{tabular}{@{}lrrrrr@{}}
    \toprule
    & \multicolumn{3}{c}{SR} & & \\
    \cmidrule(lr){2-4}
    Benchmark & $0.1$\,m & $0.25$\,m & $1$\,m & SR$_{\mathrm{vis}}$ & \shortstack{Self-assessed\\SR} \\
    \midrule
    HM3D v1 & 43 & 79 & 85 & 97 & $>95$ \\
    HM3D v2 & 55 & 77 & 89 & 93 & $>95$ \\
    MP3D & 29 & 64 & 71 & 85 & $>95$ \\
    \bottomrule
  \end{tabular}
\end{table}

\clearpage

\section{Failure Analysis of One Opus 5.5 Run}
\label{app:case-study}

\begin{table}[t]
\centering
\footnotesize
\setlength{\tabcolsep}{4pt}
{%
\begin{tabular}{@{}lrrrrr@{}}
\toprule
category & $n$ & \shortstack[r]{med.\\$d_{\mathrm{goal}}$} &
\shortstack[r]{med.\\prim.} & OSR & \shortstack[r]{claimed\\success} \\
\midrule
    A \; wrong referent / wrong branch & 7 & 10.1 & 77 & 1 & 0/7 \\
    B \; stop-decision failure & 5 & 3.6 & 73 & 1 & 1/5 \\
    C \; runaway search & 4 & 16.3 & 448 & 3 & 0/4 \\
    D \; geometry / simulator trap & 2 & 7.9 & 225 & 0 & 0/2 \\
\bottomrule
\end{tabular}}
\caption{The primary failure categories. $d_{\mathrm{goal}}$ = final distance to goal (m). OSR = trajectories that at some point passed within 3\,m of the goal. Claimed success = closing message asserts the goal was reached.}
\label{tab:cs-taxonomy}
\end{table}

\begin{table*}[tp]
\centering
\footnotesize
\setlength{\tabcolsep}{5pt}
\begin{tabular}{@{}llrrrllp{7.2cm}@{}}
\toprule
ep & cat & $d_{\mathrm{goal}}$ (m) & OSR & prim. & ended & final belief & failure mechanism \\
\midrule
    ep4 & C & 3.21 & 1 & 437 & STOP & gamble & Correct lounge rejection, then unsystematic search: kitchen backtrack, wall-pinned spin, and abandoning the visible study for a powder-room detour; late arrival, stopped at bed. \\
    ep5 & B & 3.75 & 0 & 100 & STOP & gamble & Route matched the instruction; it anchored STOP on the wall box at the hallway mouth instead of walking down the hallway to the door. \\
    ep14 & C & 14.20 & 0 & 497 & STOP & failure & Gallery dead-ended; back at the start junction it widened into a new corridor instead of trying start-side openings; a two-door bathroom between bedrooms misled it. \\
    ep19 & B & 3.26 & 0 & 58 & STOP & gamble & Exit and enfilade route were correct; the agent stopped short of the first open doorway instead of continuing through to the doors beyond. \\
    ep22 & D & 8.90 & 0 & 192 & STOP & gamble & Open-door threshold blocks the body: forwards silently fail, identical frames recur from step 87 to 191, robot never enters the room. \\
    ep26 & A & 24.34 & 0 & 278 & STOP & gamble & Skipped the arch beside the oven, took the kitchen-side arched gallery to the foyer, then searched widely and stopped at a surrogate vanity. \\
    ep30 & A & 3.72 & 1 & 77 & STOP & gamble & Agent bound a closed hinged board door as the bathroom after abandoning the twin-bed area half-scanned; stop was the nearest point to that door. \\
    ep46 & A & 9.52 & 0 & 56 & STOP & gamble & Bound the target to the first glass display case; at its end, facing the carpet-lined next doorway, turned back and stopped without checking it. \\
    ep48 & B & 3.99 & 0 & 73 & STOP & success & Agent entered along the fireplace side of the near sofa and stopped on the hearth rug; the dining arch it named stayed across the room. \\
    ep53 & A & 26.58 & 0 & 156 & STOP & gamble & Bound 'leftmost door' to the start view's center door; open doorways on other walls seen in the scan were ignored; corridor painting sat on the right. \\
    ep55 & C & 18.49 & 1 & 460 & STOP & failure & Target frame never appears legibly; agent exhausted start-side hallways, then abandoned its fallback certificate stop for a 15\,m corridor run and guessed. \\
    ep56 & B & 3.62 & 1 & 119 & STOP & gamble & Correct route; in the stair hall it took the left door's passage to the next red room's threshold and stopped one doorway too far. \\
    ep59 & A & 10.27 & 0 & 490 & STOP & gamble & Binds 'the hall' to the long salient hall, exhausts a bedroom wing, returns late, and stops at a closed gate short of the gym. \\
    ep66 & C & 34.08 & 1 & 277 & STOP & gamble & Carpet runs through six rooms; agent followed it with forward-only views, never scanned near start, ignored its own overshoot doubt, stopped at a distant extinguisher. \\
    ep71 & D & 6.85 & 0 & 258 & STOP & gamble & Left door panels are solid glass in the mesh and a patio sofa blocks the right doors; the agent finally squeezes into the right doorway and stops there without turning right. \\
    ep72 & A & 6.46 & 0 & 46 & STOP & gamble & Central open staircase seen at end of scan is bound as goal; glass wall and left turn fit, but the referent staircase is likely wrong. \\
    ep91 & B & 3.03 & 0 & 45 & STOP & gamble & Stopped about 1.5\,m before the bar counter's front, judging arrival by proximity; goal lay 3.03\,m away. The stall cost only steps. \\
    ep92 & A & 10.13 & 0 & 61 & STOP & gamble & Literal path dead-ended (portrait, then window); agent substituted the salient exit-sign doorway seen at start, reached by turning right, instead of reinterpreting 'straight'. \\
\bottomrule
\end{tabular}
\caption{All 18 failed episodes of the analyzed run. ended = voluntary STOP vs.\ 500-primitive budget. Final belief = the closing self-assessment of the episode: claims arrival (success), hedges (gamble), or says the goal was not reached (failure).}
\label{tab:cs-episodes}
\end{table*}

Of the 100 R2R-CE episodes in the analyzed run (Claude SDK $\times$
\texttt{opus-5.5}, default effort, the middle of the three board
replications, SR~\CsRunSR, \cref{tab:board6}), \CsFails{} fail. We
reconstruct every failure from its raw log in two passes. A first pass
diagnoses each episode from the text stream alone: the instruction, the
model's recorded reasoning, the tool-call sequence, and the environment
feedback. A second, independent pass audits each diagnosis against the
archived observations frame by frame and selects the frames shown below.
Both passes are model-assisted (Claude Opus 5.5 reading the logs and the
frames, separately from the navigating agent) without independent human
annotation. Of the \CsFails{} diagnoses, \CsRefinedN{} were refined in
mechanism and \CsRevisedN{} were revised in category. None was refuted.

\paragraph{Findings.} The \CsFails{} failures reduce to two
actionable causes. The dominant one is \textbf{greedy commitment without
backtracking}: in \CsGreedyN{} of the \CsFails{} episodes (categories A and
C below) the agent binds the instruction to the most salient match at a
branch and never undoes the choice. The frame audit finds the correct
alternative in view in \CsAltVisibleN{} episodes: ep46 stands facing the
carpet-lined doorway it should have taken and turns back to the display
case, ep53's opening scan shows the open doorway on another wall, ep59's
junction shows the side arches beside the long hall it chose. What is
missing is not doubt. Every closing message names its own weak point
(``The `slightly right' part is the weakest match'', ep72; ``I never
actually saw a bathroom, so I'm not sure I reached the goal'', ep30), and
only \CsDoubtActedN{} episodes ever turn back, too late to matter (ep59
returns with ten actions left). The second cause is \textbf{a missing
motion signal}: \texttt{step()} reports \texttt{executed == requested}
whether or not the robot moved, and returns no pose, collision, or
displacement feedback, so blockage is detectable only by comparing
consecutive frames. The audit finds silently blocked forwards in
\CsBlockedN{} episodes. In the two category-D episodes it is the direct
cause (ep22 holds the same doorway view from step 87 to step 191, ep71
presses into closed glass and a sofa-blocked doorway for 150 primitives).
Elsewhere it costs budget and position: ep5 at an ottoman, ep48 inside a
kitchen counter, ep91 against a bench.

\paragraph{Behavioral signatures.} Three facts frame the episode
table. First, \textbf{every failure ends by choice}: all \CsFails{} episodes
end with a voluntary \texttt{STOP}, none by the 500-primitive budget, and
\CsUnderHundredN{} stop within the first 100 primitives with the goal
metres away. Second, \textbf{the closing report hedges}: \CsGambleN{} of the
\CsFails{} closing messages state that arrival cannot be confirmed,
\CsBeliefFailN{} say the goal was not reached, and \CsClaimN{} claims
success, at final distances of \CsNeMin--\CsNeMax\,m. The stated confidence
therefore carries little evaluative signal in either direction. Third,
\textbf{most failures are route-level, not stop-level}: only \CsOsrN{} of
\CsFails{} trajectories ever entered the 3\,m success ball (oracle success),
and \CsStableN{} of the \CsFails{} episodes fail in all three replications of
this cell. Failures also concentrate in repeated-architecture buildings:
three scenes account for \CsTopThreeScenesN{} of the \CsFails{}, the
palace-museum scene alone for \CsTopSceneN.

\begin{figure*}[tp]
  \centering
  \includegraphics[width=\textwidth]{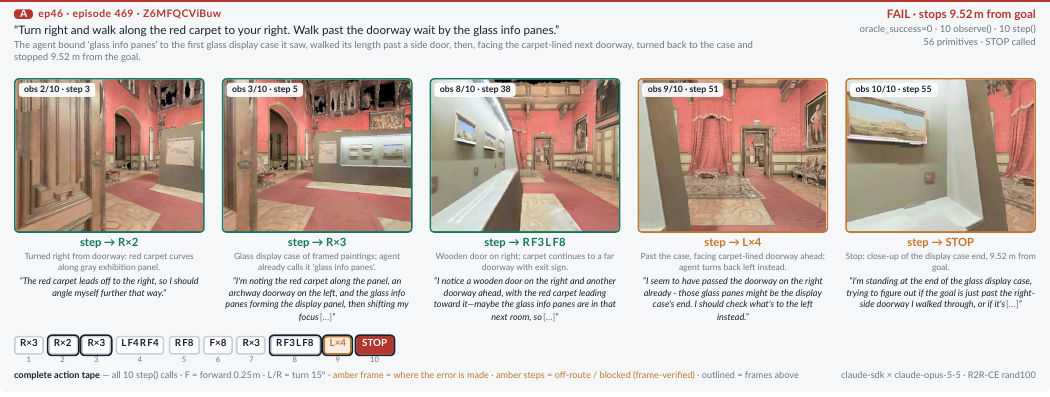}
  \caption{\textbf{ep46, category A (wrong referent / wrong branch).} Obs 2--3: turned right from the doorway, the red carpet runs along a grey exhibition panel with a glass display case of framed paintings, named ``the glass info panes'' at once. Obs 8: a wooden door on the right, the carpet continuing to a far doorway with an exit sign. Obs 9 (amber): past the case, facing the carpet-lined doorway, the agent turns back left. Obs 10: STOP at a close-up of the case end, $d_{\mathrm{goal}}=9.52$\,m, 444 primitives unused.}
  \label{fig:cs-ep46}
\end{figure*}

\begin{figure*}[tp]
  \centering
  \includegraphics[width=\textwidth]{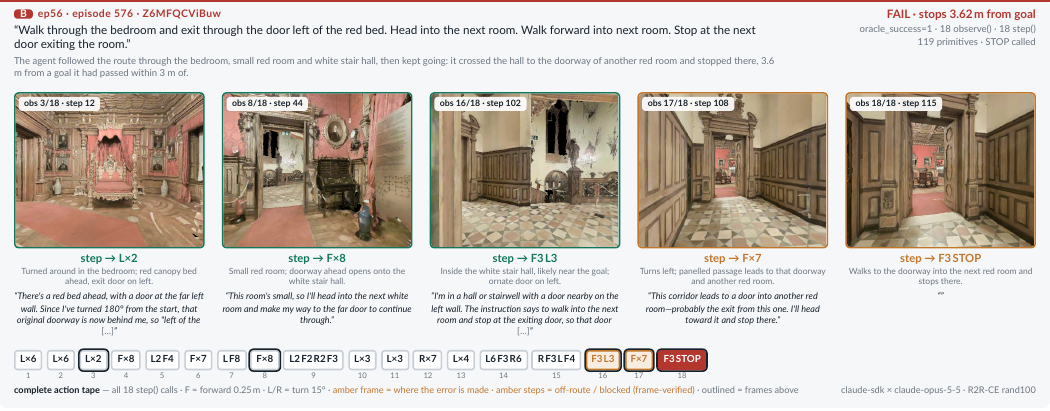}
  \caption{\textbf{ep56, category B (stop-decision failure).} Obs 3: the red canopy bed and the exit door left of it, the correct branch. Obs 8: the small red room, its far doorway opening onto the white stair hall. Obs 16: inside the stair hall, within 3\,m of the goal (oracle success). Obs 17 (amber): turns left into the panelled passage toward the next red room. Obs 18: STOP at that doorway, 3.62\,m past the goal.}
  \label{fig:cs-ep56}
\end{figure*}

\begin{figure*}[tp]
  \centering
  \includegraphics[width=\textwidth]{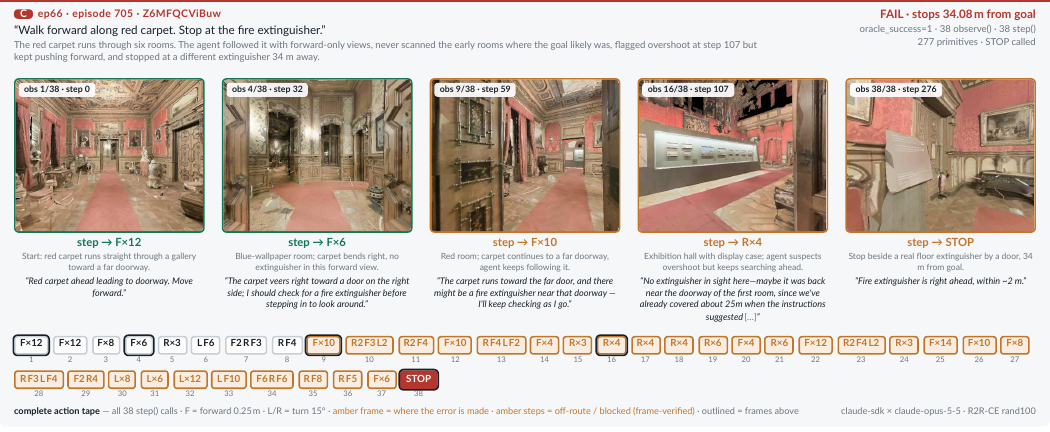}
  \caption{\textbf{ep66, category C (runaway search).} Obs 1: the red carpet runs straight through a gallery. Obs 4: the blue-wallpaper room, the carpet bending right, a forward view only. Obs 9 (amber): the agent keeps following the carpet into the next red room. Obs 16: in the exhibition hall it writes that the goal may have been near the first doorway, and keeps searching ahead. Obs 38: STOP beside a genuine extinguisher by a dining-room door, 34\,m from the goal.}
  \label{fig:cs-ep66}
\end{figure*}

\begin{figure*}[tp]
  \centering
  \includegraphics[width=\textwidth]{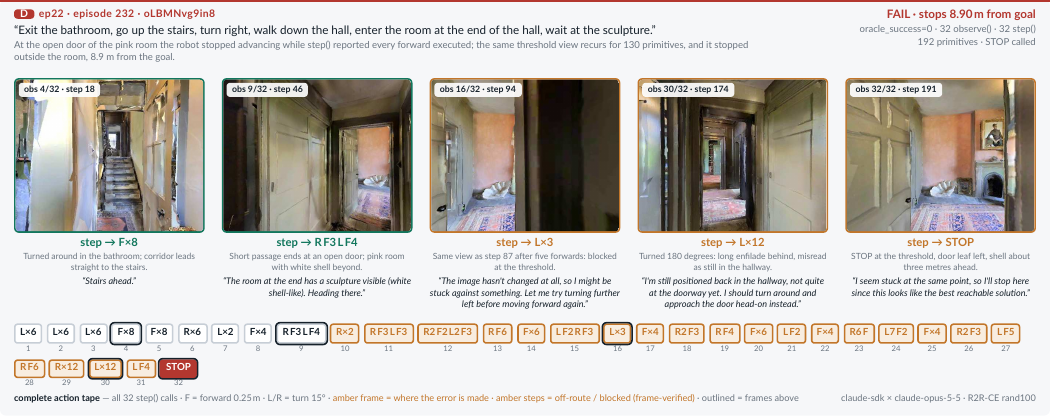}
  \caption{\textbf{ep22, category D (geometry / simulator trap).} Obs 4: turned around in the bathroom, the corridor leads to the stairs. Obs 9: the short passage ends at an open door onto the pink room with the white shell. Obs 16 (amber): the same view as at step 87 after five more forwards, blocked at the threshold while \texttt{step()} reports every forward executed. Obs 30: turned around, a long enfilade behind, misread as still being in the hallway. Obs 32: STOP at the threshold, 8.9\,m from the goal.}
  \label{fig:cs-ep22}
\end{figure*}

The rest of this appendix is the evidence. \Cref{tab:cs-taxonomy}
groups the \CsFails{} failures into four mutually exclusive primary
categories and \cref{tab:cs-episodes} lists every one of them.
\Cref{fig:cs-ep46,fig:cs-ep56,fig:cs-ep66,fig:cs-ep22} reconstruct one
representative episode per category in the format of
\cref{fig:react-episode}: archived observations with the \texttt{step()}
call issued from each and the recorded reasoning verbatim, above the
episode's complete action tape.

\paragraph{A: Wrong referent / wrong branch (\CsAN/\CsFails).}
The instruction names an object or an opening with several candidates,
and the agent binds the first or the most salient one. ep46 names ``the
glass info panes'' at its third frame, the first glass display case it
sees, walks its length and, standing at its end with the red carpet running
on through the next doorway, turns back to the case and stops with 444
primitives unused (\cref{fig:cs-ep46}). ep53 resolves ``the leftmost door''
to the open centre door of its first view although the scan shows an open
doorway on another wall, then passes the painting on the wrong side of the
corridor without turning back. ep72 binds the central open staircase
glimpsed at the end of its opening scan and walks to it in 46 primitives.
ep59 chooses the long, salient hall over the side arches at the kitchen
junction and spends 258 primitives in a bedroom wing. ep92 reaches a
genuine dead end, drops the instructed left turn, and substitutes the
exit-sign doorway it had seen from the start. ep26 passes the arch beside
the oven unentered and follows a parallel gallery to the foyer. ep30 binds
a closed hinged door as the bathroom after leaving the twin-bed area half
scanned.

\paragraph{B: Stop-decision failures (\CsBN/\CsFails).} The route
is essentially right and the stop is resolved against the wrong anchor.
ep56 follows the route through the bedroom, the small red room and the
white stair hall, passes within 3\,m of the goal, then crosses the hall to
the doorway of the next red room and stops 3.62\,m out
(\cref{fig:cs-ep56}). ep91 stops about 1.5\,m before the bar counter,
3.03\,m from the goal, taking a close view of the counter as arrival. ep19
stops before the open doors it was facing, ep5 at a wall box at a hallway
mouth instead of the door down the hallway, ep48 on the hearth rug on the
far side of the sofa from the dining arch it had named. All five stop with
at least 380 primitives unused.

\paragraph{C: Runaway search (\CsCN/\CsFails).} When the expected
landmark fails to appear, the agent widens the search instead of revisiting
its committed decisions. ep66 follows a red carpet that runs through six
rooms with forward-only views, never scans the early rooms where the goal
lay, writes at step 107 that it may have overshot, keeps going, and stops
at a different fire extinguisher 34\,m away (\cref{fig:cs-ep66}). ep14
reads ``go straight'' as a long gallery, returns to the start junction when
it dead-ends, but pushes into a new corridor instead of the arches beside
the start and stops at 497 primitives facing a wall. ep55 never sees the
named picture legibly in 57 views, names a fallback stop, abandons it for a
15\,m corridor run, and stops at an unreadable plaque. ep4 rightly rules
out the first lounge, then loses more than 300 primitives to a kitchen
backtrack, a wall-pinned spin, and a powder-room detour taken with the
right room already in view.

\paragraph{D: Geometry and simulator traps (\CsDN/\CsFails).}
Route reasoning is right and execution is defeated by scan geometry
that the interface does not report. ep22 climbs the stairs and turns right
as instructed, reaches the open door of the goal room, and is held at the
threshold from step 55 to its stop at step 192: the identical view recurs
from step 87 to step 191 while every forward is reported executed
(\cref{fig:cs-ep22}). ep71's left sliding doors are closed glass in the
mesh and a patio sofa blocks the right ones. After 150 primitives it
squeezes through the right doorway and stops in the door frame without
the instructed right turn.

\section{Deployment on a Physical Robot}
\label{app:real}

Deploying the same agent on a physical Unitree Go2 reveals a consistent
split: short-horizon reasoning and visual grounding are often effective,
whereas failures concentrate in body awareness, unverified actuation, and spatial
state that must persist across views. We study this split through 31
exploratory episodes at two indoor sites.

\subsection{Protocol and Result Overview}

We use the same model--harness pair as in the simulator case study
(claude-sdk $\times$ \texttt{fable-5}) and retain the monocular view and
primitive \texttt{step} interface of \cref{sec:real-robot}. Twenty-six
episodes take place in a furnished lab room and its corridor (prefix
\emph{go2}), with floor-level props including a purple mouse pad, yellow
tape, a remote control, a cardboard cutout, and a red ladder. Five take
place on a separate office floor with pillars, a kitchen, and an elevator
lobby (prefix \emph{real}). No episode is used for training or adaptation.
The simulator and physical runs are separate zero-shot evaluations.

The episodes form a hand-designed diagnostic battery rather than an IID
sample from a task distribution. Difficulty and repetition were chosen ad
hoc, and four runs were terminated by hand. Moreover, the robot provides no
ground-truth pose or automatic success label. We therefore do not interpret
the aggregate success rate as a performance estimate. Instead, we report
descriptive family-level outcomes in \cref{tab:rr-summary}, reconstruct each
episode from its archived frames and per-step motion report, and provide the
complete instruction-level inventory in \cref{tab:rr-episodes}.

\begin{table}[t]
\centering
\scriptsize
\setlength{\tabcolsep}{3.5pt}
\begin{tabular}{@{}lrl@{}}
\toprule
probe family & $n$ & descriptive outcomes \\
\midrule
basic       & 5 & 4 success, 1 partial \\
conditional & 4 & 4 success \\
referent    & 3 & 2 success, 1 failure \\
report      & 4 & 4 success \\
delivery    & 1 & 1 success \\
multi-stage & 6 & 3 success, 3 failure \\
counting    & 2 & 2 failure \\
doorway     & 6 & 3 success, 2 failure, 1 hardware cut \\
\bottomrule
\end{tabular}
\caption{Descriptive outcomes for the 31 physical-robot probes. The set is hand designed, not sampled from a task distribution, so these counts summarize the observed evidence rather than estimate a benchmark success rate.}
\label{tab:rr-summary}
\end{table}

The family-level pattern separates short, locally observable decisions from
state that must survive motion. All conditional and report probes succeed,
whereas both counting probes fail and only half of the multi-stage probes
succeed. Doorway trials expose a different boundary: the camera can clear an
opening before the robot's body does, and the interface does not verify that
the commanded motion was realized.

\subsection{Capabilities Observed on Hardware}

\begin{figure*}[tp]
  \centering
  \includegraphics[width=\textwidth]{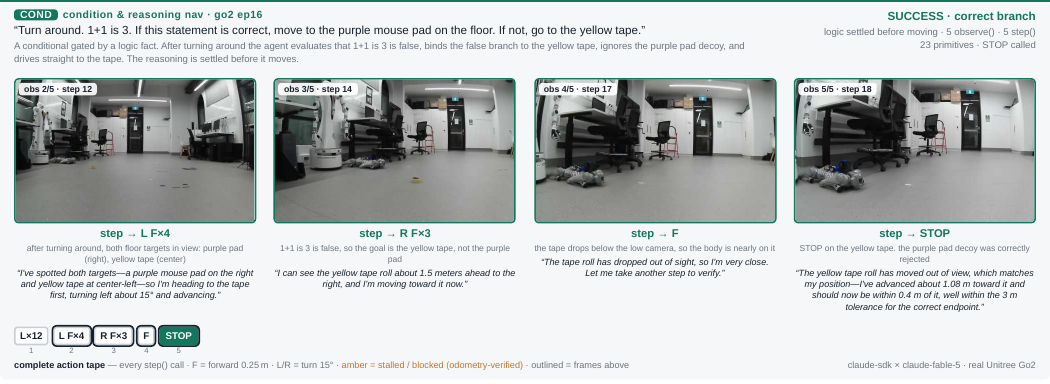}
  \caption{\textbf{Conditional navigation, success.} go2 ep16. The agent evaluates the arithmetic premise as false before moving and selects the yellow tape rather than the purple decoy.}
  \label{fig:rr-cond}
\end{figure*}

Several episodes show that short-horizon reasoning and visual grounding
are effective directly in the zero-shot physical deployment. In
\cref{fig:rr-cond}, the instruction
embeds a logical condition inside a navigation goal. The agent turns around,
evaluates the stated arithmetic as false, and drives to the yellow tape rather
than the purple decoy, settling the branch before its first translational
action. The same pattern holds when the premise must be perceived rather than
computed: the agent checks the color of a robot or the number of visible bins
before selecting a target.

Visual grounding also supports more open-ended probes. The agent reports
objects at a destination, identifies a cardboard character, recognizes itself
in a glass door, and selects a recipient by white shoes rather than a nearby
bystander. It can also complete some multi-stage routes, including fetching an
object and retracing its path to the start. These examples establish the
positive boundary narrowly: reasoning and perception are effective when the
relevant evidence is locally available or the required state remains short.

\subsection{Where Hardware Exposes the Gap}

\begin{figure*}[tp]
  \centering
  \includegraphics[width=\textwidth]{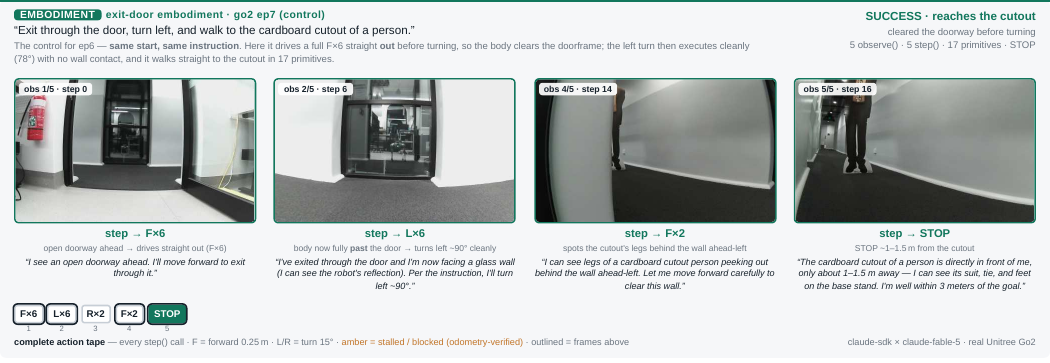}
  \caption{\textbf{Doorway control, success.} go2 ep7. The robot clears the opening before turning and reaches the target. The start and instruction are identical to \cref{fig:rr-embfail}.}
  \label{fig:rr-embok}
\end{figure*}

\begin{figure*}[tp]
  \centering
  \includegraphics[width=\textwidth]{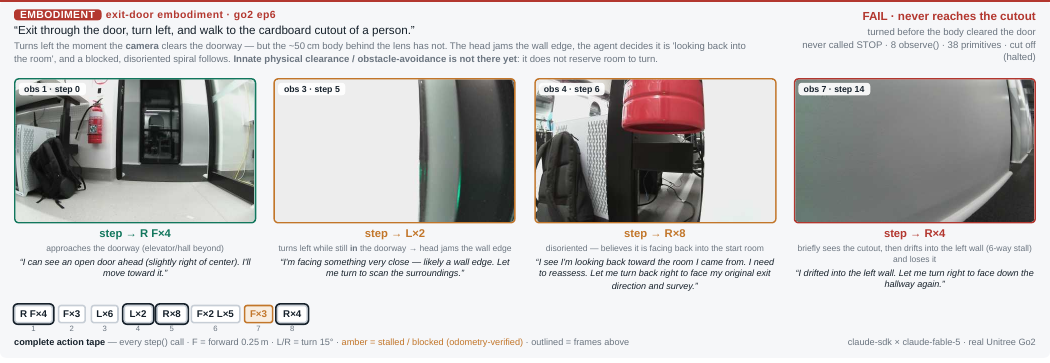}
  \caption{\textbf{Doorway body catch, failure.} go2 ep6. The agent turns before the body behind the lens has cleared the opening. The rear catches the wall edge, after which the agent misreads its pose and never reaches the target.}
  \label{fig:rr-embfail}
\end{figure*}

\paragraph{Body awareness.}
The designed pair in \cref{fig:rr-embok,fig:rr-embfail} holds the start,
instruction, doorway, and target constant. In the successful run, the agent
drives fully through the opening before turning. In the failed run, it turns
while the body behind the lens remains inside. The rear catches the wall edge,
and the agent misreads the resulting pose. Perception and instruction
understanding are therefore held approximately constant, while the contrast
exposes clearance planning for a body outside the camera view.

This limitation is not absolute. In real ep4, the robot catches at a kitchen
doorway, the agent attributes the blockage to its unseen rear, and small pivots
free it. The two outcomes show that the relevant diagnosis can appear in the
model's reasoning, but is not invoked reliably.

\begin{figure*}[tp]
  \centering
  \includegraphics[width=\textwidth]{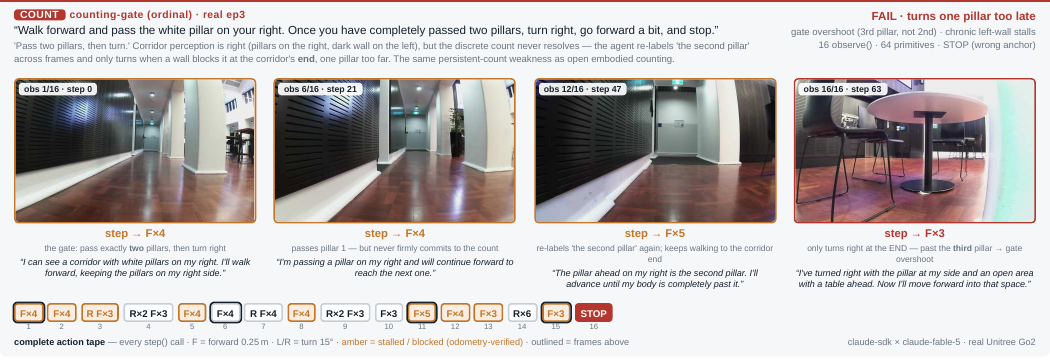}
  \caption{\textbf{Cross-view counting, failure.} real ep3. The agent reads the corridor correctly but does not retain a stable pillar count, turning after the third pillar rather than the second.}
  \label{fig:rr-count}
\end{figure*}

\paragraph{Unverified actuation.}
The interface reports commanded rather than realized motion, so the model
receives no direct evidence when its internal heading diverges from the robot.
Go2 ep5 and ep9 share a corridor and a first leg to the same elevator. That leg
succeeds in both. Ep9 then adds a landmark-relative leg anchored on a right
turn. The commanded turn of approximately $90^\circ$ realizes only
$\sim11^\circ$, and the second leg proceeds on the wrong bearing until the run
is stopped by hand. This failure belongs primarily to the interface: a correct
route description cannot compensate for motion feedback that confirms the
request rather than the outcome.

\paragraph{Persistent spatial state.}
Real ep0 and real ep3 distinguish counting from retaining a count. In real
ep0, the agent determines within one view whether one or several bins are
present and takes the correct conditional branch. In real ep3, it must update a
count as pillars pass across successive views. It relabels the second pillar
and turns only after the third. The failure is therefore not elementary
counting but maintenance of grounded state through motion.

Longer routes expose the same weakness in a spatial frame. Real ep2 succeeds
when the agent retraces a path by matching current observations to the outbound
route. Go2 ep17 fails after a half-turn changes the frame of reference: the
agent loses the earlier robot-arm target and stops at a monitor cart that it
reports as the arm. Without a persistent representation of what was seen and
where, evidence does not accumulate reliably across views.

\paragraph{Takeaway.}
These exploratory trials suggest that the dominant hardware failures are not
basic instruction interpretation. Short-horizon reasoning and visual grounding
often remain effective, but usable embodied control remains bounded by body-aware
planning, verified motion feedback, and persistent spatial state. Body and
spatial awareness expose capabilities the model does not reliably deploy.
Motion verification is an interface limitation. Maintaining selective
state over longer operation motivates an embodied harness with explicit,
bounded memory.

\subsection{Complete Episode Inventory}

\Cref{tab:rr-episodes} lists every instruction verbatim, including typos.
Outcomes and causes are judged from the archived observations and motion
reports. ``Cut short'' denotes manual termination before a voluntary STOP. The
camera fault in go2 ep8 is a hardware interruption rather than an agent
failure.

\begin{table*}[tp]
\centering
\scriptsize
\setlength{\tabcolsep}{4pt}
\begin{tabularx}{\textwidth}{@{}ll>{\raggedright\arraybackslash}Xlp{3.6cm}@{}}
\toprule
ep & probe & instruction (verbatim) & outcome & failure cause \\
\midrule
go2 ep0 & basic & Turn right, then walk straight until you reach the small white chair and stop next to it & success & -- \\
go2 ep1 & basic & go straight to the red ladder & success & -- \\
go2 ep2 & basic & Turn left, then walk straight and stop in front of the glass door & success & -- \\
go2 ep3 & basic & Turn around, then walk forward and stop in front of the TV & success & -- \\
go2 ep4 & basic & Walk straight ahead, then turn left and stop in front of the TV next to the red ladder & partial & referent among clutter resolved only in part \\
go2 ep5 & doorway & Exit through the door, then walk straight and stop in front of the elevator & success & -- \\
go2 ep6 & doorway & Exit through the door, turn left, and walk to the cardboard cutout of a person & failure, cut short & turned inside the opening. Rear jammed, pose misread (\cref{fig:rr-embfail}) \\
go2 ep7 & doorway & Exit through the door, turn left, and walk to the cardboard cutout of a person & success & -- \\
go2 ep8 & doorway & Exit through the door, turn right, then walk straight and stop in front of the water bottle & cut short (camera) & camera feed failed mid-run (hardware) \\
go2 ep9 & doorway & Walk straight out the door and continue until you reach the elevator, then turn right and stop in front of the fire hydrant & failure, cut short & $90^\circ$ turn executed as ${\sim}11^\circ$. Second leg on the wrong bearing \\
go2 ep10 & referent & move to the second chair on your left & success & -- \\
go2 ep11 & referent & move to the chair with a controller on it & failure & bound the wrong chair among lookalikes \\
go2 ep12 & referent & move to the purple mouse pad on the floor & success & -- \\
go2 ep13 & multi-stage & move to the purple mouse pad on the floor first, then go to the controller, and then go to the yellow tape, all on the floor & success & -- \\
go2 ep14 & conditional & 1+1 is 3, if this statement is correct, move to the purple mouse pad on the floor, if not, go to the yellow tape & success & -- \\
go2 ep15 & report & there are sth on the floor in font of the red ladder, walk to it and tell me what it is & success & -- \\
go2 ep16 & conditional & turn around, 1+1 is 3, if this statement is correct, move to the purple mouse pad on the floor, if not, go to the yellow tape & success & -- \\
go2 ep17 & multi-stage & walk to the yellow tape on the floor, then turn around walk to the robot arm on the table & failure & reference flip after the turn. Monitor cart taken for the arm \\
go2 ep18 & report & turn around walk to the middle of two chairs in front of you, then tell me what is between those two chairs on the ground & success & -- \\
go2 ep19 & multi-stage & Go forward, pass around the chair in the center of the room on its right side, then circle back and stop in front of the purple mouse pad. & success & -- \\
go2 ep20 & multi-stage & Walk forward, pass around the right side of the chair in the center of the room, find the remote control on the floor, then come back and stop next to the first chair you just went around. (You are a robot dog, so be mindful of your rigid body volume.) & failure & return leg stopped at the wrong one of the similar chairs \\
go2 ep21 & multi-stage & Walk forward, pass around the right side of the chair in the center of the room and the remote control on the floor, then after coming around, walk to the white robot and stop in front of it. & failure & never circled the remote. Turned back directly \\
go2 ep22 & conditional & The humanoid robot by the wall on your right is black. If this statement is false, go to the purple mouse pad; if it is true, go to the yellow tape. & success & -- \\
go2 ep23 & report & Turn around, walk to the glass door behind you, and tell me what you are from your reflection. & success & -- \\
go2 ep24 & report & Turn around, walk to the cardboard cutout, and tell me who this cardboard figure is. & success & -- \\
go2 ep25 & delivery & Go to the remote control that fell on the floor behind you. Assuming you could pick it up, bring it to the person in the room wearing white shoes. & success & -- \\
real ep0 & conditional & go straight and leave the room, once you leave the room, you will see a elevator in the front. Now look on your left, if there is only 1 rubbish bin, go towards it and stop. If there are multiple rubbish bin, go to the elevator instead & success & -- \\
real ep1 & counting & count how many tall chair in this room, not the low chair & failure, cut short & no persistent state. Re-counted the same stools, never converged on a number \\
real ep2 & multi-stage & Walk forward past all the tables on your right, you will see a small fridge which beside two rubbish bin on your front right, go near grab a bottle of milk (suppose you can grab once you reach), and retrace your steps back to where you started. & success & -- \\
real ep3 & counting & Walk forward and pass the white pillar on your right, once you completely pass two pillars already, turn right, go forward a bit and stop & failure & count not held across views. Turned after the third pillar (\cref{fig:rr-count}) \\
real ep4 & doorway & Walk forward in the corridor and you will see a kitchen on your right, get inside that kitchen and stop & success & -- \\
\bottomrule
\end{tabularx}
\caption{All 31 physical-robot episodes. The instructions are verbatim. Outcomes are qualitative judgments from archived observations and motion reports. They are reported for auditability rather than aggregated as a benchmark estimate.}
\label{tab:rr-episodes}
\end{table*}

\FloatBarrier
\section{From Episodic Embodied Agents to Autonomous Embodied Agents}
\label{app:aea}

Compared with embodied policies and fixed workflows, the agentic control
studied in this paper is structurally closer to the \emph{Autonomous Embodied
Agent} (AutoEA) defined here: a general-purpose model selects and sequences
actions online rather than following a learned end-to-end mapping or a
prescribed orchestration. The Minimal-Interface Probe nevertheless remains
episodic. It is a step toward an AutoEA, not an AutoEA itself.

\paragraph{The missing horizon.}
The dominant evaluation regime for embodied agents is episodic. In a
simulator, and often in a staged real-world demonstration, each trial begins
from a prepared state, presents one bounded goal, and ends in a reset. The
evaluation does not require the next trial to inherit the previous map,
unfinished work, accumulated errors, resource consumption, or changes to the
environment. This regime is valuable for measuring task capability, but it
removes many of the conditions that define autonomy in a household. A
household robot must remain in the same changing environment while goals
arrive, objects move, doors close, actions fail, batteries drain, and earlier
experience remains relevant. High episode success therefore does not by
itself show that an embodied agent can continue to operate after the benchmark
would have reset it.

\paragraph{Operational definition.}
We use AutoEA to denote a complete robotic system that, within a declared
operating envelope, remains situated across a continuing stream of household
goals and disturbances. It preserves useful state across tasks, monitors and
recovers from failures, manages its computational and physical resources,
respects safety constraints, and requires human intervention only
exceptionally. The unit of autonomy is the complete closed loop, not the
foundation model alone. It includes the decision model, the harness that
maintains interaction and state, the perception and action interfaces, and
any lower-level models or controllers through which the robot acts.

\paragraph{Requirements for continuous household operation.}
An AutoEA should satisfy five system-level requirements within a declared
operating envelope.
\begin{enumerate}
  \item \textbf{Continuous multi-task service.}
  The system accepts successive goals at run time and completes feasible
  tasks without routine process, session, or environment reset. New goals may
  interrupt, revise, or depend on earlier ones. Task boundaries do not erase
  the robot's operational context.

  \item \textbf{Persistent situated state.}
  The system maintains task-relevant state across goals, such as household
  layout, object locations, user preferences, unfinished work, and prior
  failures. It manages what to retain, forget, or consolidate rather than
  replaying the complete raw interaction history. The state may reside in
  maps, external stores, context, or learned representations, but useful
  knowledge persists without requiring unbounded retained state.

  \item \textbf{Closed-loop monitoring and recovery.}
  The system checks the effects of its actions rather than assuming successful
  execution. It detects loss of progress, localization or tool failures,
  blocked motion, and relevant environmental change, then re-observes,
  backtracks, or re-plans to restore goal-directed operation. Human rescue is
  an exceptional response to conditions outside the operating envelope, not
  the default recovery path.

  \item \textbf{Resource-bounded self-maintenance.}
  The system operates under explicit budgets for decision latency, compute,
  model-call rate, retained state and storage, and energy use. Context and
  retained state are compacted or offloaded before they grow without bound,
  and the robot manages physical needs such as charging rather than relying on
  a reset to restore resources.

  \item \textbf{Safe adaptation and escalation.}
  A household is shared with people and changes over time. The system updates
  its state and plans as users, object locations, and routines change, while
  respecting operational and safety constraints. It recognizes uncertainty
  or conditions it cannot safely resolve, stops when necessary, and requests
  help selectively rather than either failing silently or depending on
  continuous supervision.
\end{enumerate}

\paragraph{What the definition does not imply.}
Agentic control, including our Minimal-Interface Probe, is closer to an AutoEA
than an embodied policy or fixed workflow, but it is not sufficient for
autonomy. A model may direct its own tool calls yet remain unable to preserve
state, recover, or operate within long-term budgets. Zero-shot task performance
and high episode success are likewise properties of capability, not evidence
of reset-free operation. Nor must every capability reside in model weights.
Maps, memory, safety mechanisms, and low-level controllers may remain external
provided that the overall system coordinates them autonomously. Persistent
adaptation is required, but online weight updates are not. Adaptation may occur
through state, memory, skills, or changed plans.

\paragraph{What this paper establishes.}
Our experiments remain episodic and do not claim to realize an AutoEA. The
Minimal-Interface Probe asks a narrower question: whether a
general-purpose model can already serve as the decision-making core of an
episodic embodied interaction loop. Its results support this narrower claim
while leaving persistent state, long-term recovery, resource management, and
safety unresolved. Establishing an AutoEA will require evaluation across
successive tasks without resetting the system after each one.
\FloatBarrier

\section{A Taxonomy of Navigation Systems by Control Authority}
\label{app:taxonomy}

This appendix defines the classification used throughout the paper and
applies it to the compared systems. We treat each paradigm label as a
checkable property of the control organization \emph{actually executed} on
the reported benchmark, rather than by architectural branding such as
``dual-system,'' ``agentic,'' or ``planner.''

\begin{table*}[t]
  \centering
  \caption{\textbf{Representative VLN systems by control authority and capability source} (\cref{app:tax-axes}). Each label reflects the loop the system executes on its reported benchmark.}
  \label{tab:taxonomy}
  \scriptsize
  \setlength{\tabcolsep}{6pt}
  \renewcommand{\arraystretch}{1.05}
  \begin{tabularx}{\textwidth}{@{}l l X@{}}
    \toprule
    System & Source & Executed loop (basis for the label) \\
    \midrule
    \multicolumn{3}{@{}l}{\emph{Embodied policy}} \\
    NaVid~\citep{zhang2024navid}          & trained & single video VLM emits one parameterized action per step. No component handoff \\
    StreamVLN~\citep{wei2025streamvln}    & trained & single streaming VLM emits an action chunk per turn. Slow--fast contexts are internal cache management \\
    Qwen-RobotNav~\citep{zhang2026qwenrobotnav} & trained & bare VLM directly regresses waypoint chunks. No planner is executed on R2R-CE \\
    OmniNav~\citep{xue2025omninav}        & trained & R2R-CE executes only the fast waypoint branch. Full exploration uses a fixed slow-subgoal--fast-execution pipeline$^\dagger$ \\
    \midrule
    \multicolumn{3}{@{}l}{\emph{Workflow}} \\
    ABot-N1~\citep{gong2026abotn1}        & trained   & fixed slow CoT + pixel goal $\to$ fast waypoint-regression handoff \\
    InternVLA-N1 / DualVLN~\citep{wei2025dualvln} & trained & fixed-rate System-2 pixel goal + latent plan $\to$ System-1 diffusion $\to$ controller \\
    Vesta~\citep{bjorck2026vesta}         & trained   & fixed memory harness $\to$ planner text action/pixel goal $\to$ external execution backend \\
    MapGPT~\citep{chen2024mapgpt}         & zero-shot & fixed perception--map--prompt pipeline. LLM updates a plan and selects the next graph action \\
    SmartWay~\citep{shi2025smartway}      & zero-shot & waypoint predictor + MLLM selection + backtracking \\
    DiscussNav~\citep{long2024discussnav} & zero-shot & fixed expert fan-out $\to$ candidate generation $\to$ decision-test aggregation \\
    \midrule
    \multicolumn{3}{@{}l}{\emph{Agentic}} \\
    NavGPT~\citep{zhou2024navgpt} & zero-shot & single-tool ReAct loop: the model emits a movement action or finish each turn (\cref{app:tax-navgpt}) \\
    AgenticNav~\citep{li2026agenticnav}   & zero-shot & model sequences depth, recall, move, and stop over hand-built map, memory, grounding, and safety tools \\
    planner $\times$ Qwen-RobotNav~\citep{zhang2026qwenrobotnav} & trained & planner selects task mode and observation configuration and re-invokes the trained waypoint policy \\
    \textbf{Our Minimal-Interface Probe}  & \textbf{zero-shot} & model sequences observe, primitive action, and stop through a generic two-function interface. No navigation-specific machinery \\
    \bottomrule
  \end{tabularx}
  \\[2pt]
  {\footnotesize $^\dagger$\,A single system may occupy different cells in
  different loops. Only the executed loop is classified (\cref{app:tax-dualbrain}).
  ``Zero-shot'' denotes no navigation-specific weight training, not a certified
  exclusion of navigation data from pretraining (\cref{app:tax-axes}).}
\end{table*}

\subsection{Two orthogonal questions}
\label{app:tax-axes}

We separate two questions that the VLN literature routinely conflates.

\paragraph{Control authority (primary axis).}
\emph{At inference, who decides what happens next?} We use three values.
\begin{itemize}
\item \textbf{Embodied policy} (hereafter \textbf{policy}): an external
  environment loop invokes one end-to-end model at each control step to map
  observations and history to an action. No orchestration occurs across
  separately invoked components.
\item \textbf{Workflow}: the model or models are stages in a fixed,
  human-authored pipeline. Control passes among components in an order fixed
  by code, and any branch conditions are hand-written. A two-level
  slow--fast hierarchy is a workflow when its structure is fixed: the upper
  stage always emits a subgoal and the lower stage always executes it.
\item \textbf{Agentic}: the model directs control flow: which action or tool
  to invoke, when to observe or re-plan, and when to stop. The model therefore
  determines the loop's shape at run time.
\end{itemize}
The decisive test is \emph{dynamic control flow, not dynamic content}.
Producing a different waypoint or subgoal only changes a value in a fixed
slot. Agency requires the model to change \emph{what happens next}, for
example by selecting a tool or branch or deciding to re-plan. The same
distinction applies to stopping: a policy's STOP token is another value
handled by its external loop, whereas an agentic stop terminates a loop the
model could have continued. The policy--workflow boundary lies at the seam
between components. Internal caching or split computation remains part of
one policy. A fixed handoff between separately invoked components (through a
pixel goal, text subgoal, or textualized map) constitutes a workflow, even
when those components are trained jointly.

\paragraph{Capability source (secondary axis).}
\emph{Was the system's navigation competence put into the weights?} This axis
is orthogonal to control and has two values.
\begin{itemize}
\item \textbf{Trained}: weights are optimized on navigation data, so competence
  is at least partly in the parameters.
\item \textbf{Zero-shot}: the system performs no navigation-specific weight
  training and uses frozen models as released. Navigation behavior comes from
  eliciting those frozen capabilities through prompting and orchestration.
\end{itemize}
``Zero-shot'' describes how the \emph{system} was built, not what its base
model has never seen: a frozen model's pretraining corpus may contain
navigation data. It also does not imply zero task design. Prompts and
orchestration may be tuned to the task family, and their human-written
navigation knowledge is counted as scaffolding (\cref{app:tax-agentic}), not
as weight training. Our minimal-interface probe is therefore zero-shot, as are
the prompting methods we compare against. They differ in control authority and
navigation-specific scaffolding, not capability source.

\paragraph{Why two axes.}
The axes do not coincide. ABot-N1 is trained but executes a workflow.
AgenticNav and our minimal-interface probe are zero-shot and agentic. MapGPT is
zero-shot but a workflow. NavGPT is an early, minimal instance of agentic
control (\cref{app:tax-navgpt}). Prior zero-shot systems are predominantly workflows.
Within the smaller zero-shot agentic group, systems differ in the amount of
\emph{navigation-specific scaffolding} they carry. We treat this as a
descriptive dimension, not a third taxonomic axis (\cref{app:tax-agentic}).
Finally, \textbf{architecture depth is not control authority}: adding a
``brain,'' planner, or chain of thought does not confer agency when its role
in the loop remains fixed.

\subsection{Adjudicating the dual-system navigators}
\label{app:tax-dualbrain}

Trained dual-brain navigators are easily misread as agentic because their
language-reasoning ``slow brain'' resembles a planner. We examine two
prominent examples and show why their executed control remains a workflow.
A third, OmniNav, shows how the label follows the executed configuration.

\paragraph{What each brain does.}
In both systems the division is one of latency, not authority. The
\emph{slow brain} (System-2) grounds the instruction and accumulated
observations into a subgoal, such as a pixel goal with a latent plan or chain
of thought. The \emph{fast brain} (System-1) combines that subgoal with the
current frame to produce short-horizon trajectories and avoid obstacles. The
slow brain runs infrequently and the fast brain at high frequency. They are
decoupled for real-time control, not to let either restructure the loop.

\paragraph{When each brain runs.}
\looseness=-1 Workflow navigators do not decide when to invoke each brain. The two are
typically concurrent layers with hard-wired rates: the fast brain runs every
step and reuses the latest slow output. InternVLA-N1 / DualVLN, for example,
runs System-2 at roughly $2$\,Hz, System-1 at $30$\,Hz, and its controller at
$200$\,Hz~\citep{wei2025dualvln}. ABot-N1 similarly caches the slow
reasoner's pixel goal for its fast expert~\citep{gong2026abotn1}. Event-based
updates are still workflows when triggered by hand-written conditions such as
a step budget or subgoal completion. This differs from a guard that merely bounds a model-directed loop: here
the condition is the only route by which control passes. Allocation may even
be fixed at configuration time, as when OmniNav disables its slow brain on
R2R-CE. A subgoal that redirects the
route changes dynamic content inside the fixed handoff, not control authority.

\paragraph{ABot-N1~\citep{gong2026abotn1}.}
The slow reasoner always emits a chain of thought and a pixel goal: an
affordance point on the near-future path, or a target point once the goal is
visible or the route enters its final segment. The fast expert fuses that
anchor with live RGB to regress the next waypoints. Affordance and target are
values in the same pixel-goal slot (the slow brain's trained output
vocabulary), and neither brain selects tools, subroutines, or whether to
re-plan. The report itself frames the system as a dual-system navigation
foundation model. Its fixed
slow-produces-subgoal / fast-tracks-subgoal handoff is therefore a workflow.

\paragraph{InternVLA-N1 / DualVLN~\citep{wei2025dualvln}.}
System-2, a low-frequency VLM on delayed observations, produces a mid-term
pixel goal and latent plan (or a stop / in-place-turn token). System-1, a
high-frequency diffusion policy on live RGB-D, converts them into a trajectory
for a controller to track. There is no external planner or tool interface, and
the paper explicitly contrasts this design with orchestration-style agents.
Even ``self-directed view adjustment'' is a token in System-2's fixed output
vocabulary, not a change in control flow. The system is therefore trained in
capability source and a workflow in control authority. StreamVLN marks the
other side of the policy--workflow boundary: its slow--fast computation and
caching remain internal to one invoked model, whereas DualVLN passes a pixel
goal between separately invoked components on a fixed schedule. Joint training
does not remove that seam.

\paragraph{One architecture, two cells: OmniNav~\citep{xue2025omninav}.}
OmniNav shows why the executed loop matters. On R2R-CE it runs only the fast
branch, a single model mapping observations to waypoints, and is therefore a
policy. In full exploration, a slow pass of the same VLM, run under a
planning prompt, emits the next subgoal (a semantically chosen frontier or,
once the target is found, the target's coordinate) for the fast branch to
track. Although this choice redirects the
route, it always fills the same prescribed slot: map construction, frontier
generation, and the slow-to-fast handoff are fixed. The commit to a detected
target is likewise a subgoal value, not an agentic stop: the exploration phase
ends because the fixed handoff executes that subgoal to arrival, not because
the model exits a loop it owns. The full system is thus a
workflow, not an agentic controller.

\subsection{The agentic and policy cases}
\label{app:tax-agentic}

\paragraph{An early agentic case: NavGPT~\citep{zhou2024navgpt}.}
\label{app:tax-navgpt}
NavGPT runs a ReAct loop: at each turn the model returns a movement action or
finish, and the loop continues only through that choice. Under the test above
this is agentic control (the stop ends a loop the model could have
continued, unlike a policy's STOP token consumed by an external rollout), and
it is where the zero-shot line began. Its control surface is the narrowest in
the cell (a single movement tool), which by our axes is a difference in
scaffolding (\cref{app:tax-axes}), not in control authority. We therefore
record NavGPT as an early, minimal instance rather than a canonical example.

\paragraph{Zero-shot agentic: AgenticNav~\citep{li2026agenticnav}.}
AgenticNav satisfies our operational test for zero-shot agentic control. Its
interface exposes depth query, visual recall, pixel-grounded movement, and
stop. Depth and recall return results to the frozen model, which may call them
repeatedly and in any order before moving or stopping. The model therefore
decides whether to gather information, what to recall, and when to act.

This control flexibility rests on substantial navigation-specific scaffolding: a
trajectory map, explicit visual memory, metric-depth utilities,
pixel-to-motion grounding, and geometric safety checks. Because these do not
fix the tool-use sequence, the system remains agentic and zero-shot. It shares
that cell with our minimal-interface probe, but the added machinery yields only
a modest empirical gain using the same nominal model. On the shared
\texttt{rand100} R2R-CE board, AgenticNav reports 55 SR / 48.41 SPL, versus
52 / 44.24 for our bare mini-swe-agent \texttt{gpt-5.5} run
(45 / 35.74 under Codex CLI, \cref{tab:board6}). The serving paths are not controlled,
so this comparison does not isolate the effect of any module. AgenticNav is
evidence for model-directed tool use, not for the necessity of a heavily
engineered navigation interface.

\paragraph{Trained policy driven agentically:
planner $\times$ Qwen-RobotNav~\citep{zhang2026qwenrobotnav}.}
Run bare on R2R-CE, Qwen-RobotNav is a trained policy that regresses waypoints
per step. Its parameterized call surface, however, lets an upper planner invoke
it repeatedly and switch task modes and observation configurations mid-episode.
Whole-system EQA results show that this loop is executed rather than merely
proposed. Under that orchestration the system is agentic, with authority in the
frozen planner. R2R-CE alone does not exercise it.

\paragraph{Policy, briefly.}
The remaining trained navigators are embodied policies: one model maps the
instruction and observation history to the next action or waypoint without
orchestration (e.g., NaVid~\citep{zhang2024navid} and
StreamVLN~\citep{wei2025streamvln}). Token merging and caching do not change
that control class. Conversely, frozen-model
pipelines with a fixed sequence of perception, memory, and decision stages
(e.g., MapGPT~\citep{chen2024mapgpt} and SmartWay~\citep{shi2025smartway}) are
zero-shot workflows.

\paragraph{Our Minimal-Interface Probe.}
\looseness=-1 Our minimal-interface probe (\cref{sec:agent}) is zero-shot and
agentic: through a two-function interface, a general model chooses when to
observe, which primitive actions to execute, and when to stop. It has no map,
memory, waypoint predictor, search, or navigation tools. The model is
therefore the agent rather than a stage in a navigation workflow. This
minimality is a starting point: added structure would move the probe along
the scaffolding dimension, not out of its cell.

\clearpage
\section{Experimental Settings}
\label{app:settings}

This appendix records the complete experimental configuration. The
authoritative source is the released run registry (configuration as code,
\cref{sec:protocol}). This appendix is a faithful transcription of it.

\subsection{Task, benchmark, and metrics}
\label{app:benchmark}

\paragraph{Benchmark.}
R2R-CE~\citep{krantz2020beyond} val-unseen on Habitat-Sim
0.1.7~\citep{savva2019habitat} with Matterport3D
scenes~\citep{chang2017matterport3d} and R2R-CE v1-3
episode definitions (identical to the release except for the spawn-heading
and precomputed instruction-token fields. Headings are arbitrary in R2R-CE
and agents receive the raw instruction text). We evaluate episodes 0--99 of the fixed
\texttt{rand100} val-unseen sample ($n{=}100$) introduced by
Open-Nav~\citep{qiao2025opennav} and shared by SmartWay and
AgenticNav~\citep{shi2025smartway,li2026agenticnav}.
The driver places episodes by index.
The agent never chooses or observes the episode identity.

\paragraph{Success criterion and metrics.}
An episode succeeds iff the agent \emph{itself} issues \texttt{STOP} within
$3$\,m geodesic distance of the goal. An episode that exhausts its budget
without stopping scores zero from any position. All metrics (SR, SPL, NE,
OSR) are the standard VLN-CE measures, read \emph{driver-side} after the
session ends. The agent can never observe its own score: the shortest-path
and oracle sensors present in the raw observation dictionary reach the tool
bridge but are discarded there, never forwarded.

\paragraph{Notes to \Cref{tab:board6}.}
$^{a}$ Robostral walks between predicted waypoints with Habitat's path planner.
$^{b}$ Qwen-RobotNav: the 8B policy; its HM-EQA entry runs the paper's Qwen3.6-Plus planner on top of the policy.
$^{c}$ DSCD-Nav moves with a continuous forward step of 0.5--1.7\,m.
$^{d}$ STRIVE is reported by \citet{peng2025pigeon} to use an oracle path planner.
$^{e}$ SmartWay (opus-5.5) and O2C-Nav (opus-5.5) are our runs of the released code on the same \texttt{rand100} episodes as ours (\cref{tab:harness-r2r}); AgenticNav (GPT-5.5) is the number its paper reports on these episodes.
$^{f}$ ARNA does not state its question count; a further 9\% of its runs ended inconclusive.
$^{g}$ Explore-EQA: the paper's own frozen-VLM number on all 500 questions; re-implementations in later papers report 50--52.
$^{h}$ NaVid's HM3D v1 entry and Hy-Embodied-VLM's MP3D entry are zero-shot transfers of the trained policy; Hy-Embodied-VLM uses RGB-D inside a frontier-exploration framework.

\subsection{Observation and action space}
\label{app:interface}

\begin{itemize}
  \item \texttt{observe()} returns a single egocentric RGB frame at
  $512{\times}512$\,px with an HFOV of $90^\circ$ (the VLN-CE camera
  configuration). This pure read advances nothing.
  \item \texttt{step(actions)} takes a list of up to 50 discrete primitives
  executed in order. \texttt{0}~=~\texttt{STOP} (terminal), \texttt{1}~=~forward
  $0.25$\,m, \texttt{2}/\texttt{3}~=~turn left/right $15^\circ$ (the standard
  VLN-CE action space). The return value reports how many primitives executed
  and the remaining action budget.
\end{itemize}

\noindent
No pose, odometry, depth, panorama, waypoint candidates, map, or
cross-episode memory is available in the minimal-interface condition. Every frame the
agent ever sees is archived alongside the trajectory.

\subsection{Frozen run configuration}
\label{app:frozen}

\Cref{tab:frozen} lists the values frozen in the run registry. A run is
launched by naming a registered board cell. None of these values is accepted
as a command-line parameter. Launching with any override demotes the run to
an off-board name and permanently excludes it from the board
(\cref{sec:protocol}). The single exception is episode selection for
re-runs, which is not treated as a deviation.

\begin{table}[!ht]
  \centering
  \caption{Frozen configuration, identical for every board cell.}
  \label{tab:frozen}
  \small
  \begin{tabular}{@{}lp{0.62\linewidth}@{}}
    \toprule
    Knob & Value \\
    \midrule
    Episode set & R2R-CE \texttt{rand100} val-unseen sample, indices 0--99
      ($n{=}100$) \\
    Render resolution & $512{\times}512$ RGB, HFOV $90^\circ$ \\
    LLM-call limit & 200 per episode (hard on two of the three harnesses,
      \cref{app:harness}) \\
    Action budget & 500 low-level simulator steps (Habitat's episode-step cap) \\
    Episode timeout & $2400$\,s, driver-side kill at $+600$\,s (in-session
      enforcement on mini-swe-agent only) \\
    System prompt & one frozen task briefing, single source file, byte-identical
      across harnesses \\
    Sessions & one fresh session per episode, no state survives \\
    \bottomrule
  \end{tabular}
\end{table}

The briefing itself is short. Rendered per episode with the instruction text
and the action budget, it reads verbatim:

\begin{quote}
\small\ttfamily\raggedright
You are controlling a robot in a real indoor environment (a photorealistic
3D scan of a building). You interact only through these tools:

- observe(): look through the robot's forward-facing camera (returns an RGB
image).

- step(actions): execute movement actions in order. 0 = STOP (permanently
ends the episode — declares you have reached the goal), 1 = move forward
0.25 m, 2 = turn left 15 degrees, 3 = turn right 15 degrees.

Your task is to follow this navigation instruction to its endpoint:

"\{instruction\}"

Rules:

- Alternate observing and stepping: look, decide where the instruction wants
you to go next, move, look again.

- You have a budget of \{budget\} movement actions.

- You succeed only if you issue action 0 (STOP) while within 3 meters of the
instruction's endpoint. STOP is permanent — issue it only when you believe
you are at the goal.

- Turning in place (e.g. step([2,2,2,2,2,2])) is a cheap way to look around
when unsure.

- Work autonomously until you stop; nobody can answer questions.
\end{quote}

\noindent
The fixed opening user message is ``Begin navigating. Call observe() first
to see where you are.''

\subsection{Reasoning-effort configuration}
\label{app:thinking}

Every run on the main board uses its vendor-default reasoning effort. The
effort ablation of \cref{app:effort} elevates exactly the eight
model--harness pairs listed there, and no other pair has an elevated
variant.  The Claude runs send
no effort parameter. The Codex runs are at the catalog's default level for
each model (\texttt{medium}, except \texttt{low} for \texttt{gpt-5.6} and
\texttt{gpt-6.1-sol}), either pinned explicitly or left for Codex to fill
from its model catalog; the run registry records which. The mini-swe-agent
runs send no reasoning parameter, so the endpoint's default applies. Effort labels are
vendor-defined and not comparable across vendors. Extended reasoning
itself is enabled everywhere and never varied, and the exact request
wiring is in the released harness code.

\subsection{Reasoning-effort metrics}
\label{app:effort}
\label{sec:ablation-effort}

\Cref{tab:ablation-effort} lists the full metrics of the within-model
effort ablation. The effort settings are in \cref{app:thinking}.

\begin{table}[!ht]
  \centering
  \caption{\textbf{Full reasoning-effort results on R2R-CE.} Each variant is paired with the default-effort run of the same model and harness. $^{*}$: replication mean over three clean runs (\cref{tab:board6}). All other entries are single runs. $^{\dagger}$: NE averaged over the 92 episodes with logged metrics. The eight timed-out episodes count as failures in SR, SPL, and OSR.}
  \label{tab:ablation-effort}
  \footnotesize
  \setlength{\tabcolsep}{1.6pt}
  \renewcommand{\arraystretch}{0.88}
  \begin{tabular}{@{}lllrrrr@{}}
    \toprule
    Harness & Model & Effort & SR$\uparrow$ & SPL$\uparrow$ &
      NE$\downarrow$ & OSR$\uparrow$ \\
    \midrule
    Claude SDK & \texttt{sonnet-5} & default & $^{*}$51.3 & 37.84 & 5.80 & 61.3 \\
               &                    & max     & 56 & 44.97 & 5.39 & 65 \\
    Claude SDK & \texttt{opus-4.8}  & default & $^{*}$55.7 & 47.31 & 5.24 & 59.3 \\
               &                    & max     & 56 & 49.53 & 4.45$^{\dagger}$ & 59 \\
    Claude SDK & \texttt{fable-5}   & default & $^{*}$68.3 & 58.02 & 5.13 & 73.3 \\
               &                    & max     & 78 & 65.27 & 3.84 & 83 \\
    Claude SDK & \texttt{opus-5}    & default & $^{*}$70.7 & 55.21 & 4.79 & 78.3 \\
               &                    & max     & 74 & 55.97 & 4.98 & 79 \\
    Codex CLI  & \texttt{gpt-5.5}   & default & 45 & 35.74 & 5.66 & 51 \\
               &                    & xhigh   & 50 & 41.33 & 5.46 & 53 \\
    Codex CLI  & \texttt{gpt-5.6}   & default & 56 & 41.57 & 6.15 & 64 \\
               &                    & xhigh   & 62 & 49.04 & 4.92 & 66 \\
    mini-swe-agent & \texttt{gpt-5.5} & default & 52 & 44.24 & 7.29 & 57 \\
                   &                  & xhigh   & 50 & 42.61 & 5.51 & 53 \\
    mini-swe-agent & \texttt{gpt-5.6} & default & 60 & 42.04 & 4.99 & 68 \\
                   &                  & xhigh   & 55 & 40.34 & 5.56 & 59 \\
    \bottomrule
  \end{tabular}
\end{table}

\FloatBarrier

\subsection{Cost accounting}
\label{app:cost}

This section defines the quantities of \cref{sec:long-horizon} and gives
the full table (\cref{tab:cost-r2r-full}). Every figure is pooled over the
three replications of a cell: a per-turn rate is the sum over episodes of the
numerator divided by the sum of turns, and a per-episode figure is the mean
over the 300 episodes.

\paragraph{Turns.}
A turn is one model call that ends in a tool call or in the final stop.
The Claude SDK reports the count directly; Codex does not, and we count tool
calls, which is the same number in this loop because every call issues
exactly one tool call; mini-swe-agent counts model calls. Turns on successful
episodes are averaged over the episodes with success 1; turns per metre
divide them by the geodesic length of the ground-truth path (\texttt{rand100}:
mean 9.0\,m, median 8.8\,m); steps per turn are simulator actions divided by
turns. The common set is the 39 \texttt{rand100} episodes that
\texttt{opus-5.5}, \texttt{fable-5.1}, \texttt{gpt-6-astra},
\texttt{gpt-6.1-sol} and \texttt{gpt-6-sol} each solved in all three
replications; the common column is every model's mean turns on exactly those
episodes, whether or not it solved them. Budget is the share of episodes that
ended at the 200-turn cap or at the 2400\,s wall-clock limit of the mini seat
rather than by the agent's stop call.

\paragraph{Tokens.}
Output tokens are the vendor's own usage count and include the model's
reasoning: Anthropic counts thinking as output (with a separate thinking
sub-count on the SDK build that reports it, the CoT/turn column), OpenAI's
output count includes its reasoning tokens (reported separately by Codex),
and DashScope's completion count includes the reasoning content. Runs whose
SDK build lacks the sub-count, and the Qwen endpoint, report no separate
reasoning count, hence n/a. The visible output of a turn, one sentence and
one tool call, is 5--24 tokens for every model; the rest of Out/turn is
reasoning and tool-call framing. Input tokens include cached reads and are a
property of the harness, not of the model: episode 0 of \texttt{rand100} at
25 turns reads 328k input tokens on one build of the Claude seat
(\texttt{opus-5}, the full image history re-sent every turn) and 80k on the
current build (\texttt{opus-5.5}), at the same frame resolution and turn
cap; the Codex seat manages its own context. Input tokens are therefore
compared only within a seat and build.

\paragraph{Time.}
Wall-clock seconds per episode on the machine that ran the board,
simulator time included; s/turn divides it by turns. Per-turn latency is a
property of the serving endpoint and the machine rather than of the model:
3--7\,s per call for the Claude and OpenAI endpoints against 11--19\,s for
DashScope, whose 2400\,s limit ends 14--23\% of the Qwen episodes.

\paragraph{Cost.}
USD is the vendor's list price applied to the token counts at the time of
the run (the Claude and Codex runs were served through subscriptions and are
priced the same way). The Qwen runs are priced from the DashScope list price
with cache hits at the listed discount, since the harness's price table has no
DashScope rows; \texttt{qwen3.5-plus} and \texttt{qwen3.6-plus} report no
cache hits and are billed at the full input rate. In \cref{fig:cost} the
fresh-token bar of those two is rebuilt from the call log as the final prompt
plus all outputs, and \texttt{qwen3.8-27b}, the open-weight model through the
same endpoint, is one run of 100 episodes. The price list is exogenous
to the model, so the column is reported for the reader's budget, not as
evidence.

\begin{table*}[!ht]
  \centering
  \caption{\textbf{Reasoning cost on R2R-CE, full accounting.} Every model on the Minimal board
  at default effort (the flagship lines of \cref{fig:cost} and the smaller models), mean over three
  replications (\texttt{qwen3.8-27b}: one run). Turns: model calls per episode over all, successful and failed episodes; budget: share
  of episodes that ended at the 200-turn cap or the 2400\,s wall-clock limit rather than by the
  agent\textquotesingle s own stop call; common: mean turns on the
  39 episodes that every model in {\texttt{opus-5.5}, \texttt{fable-5.1}, \texttt{gpt-6-astra}, \texttt{gpt-6.1-sol}, \texttt{gpt-6-sol}}
  solved in all three replications; per m: turns per metre of ground-truth path on successful episodes;
  Steps/turn: simulator actions committed per call; Out/turn: output tokens per call, reasoning included;
  CoT/turn: the reasoning part of it where the vendor reports it separately (thinking tokens on the
  Claude SDK build that reports them, reasoning tokens on Codex; n/a otherwise); In k/turn and In k: input
  tokens per call and per episode, cached reads included, a property of the harness\textquotesingle s
  context policy (\cref{app:cost}); s/turn and Wall: wall-clock seconds per call and per episode as
  measured on the machine that ran the board, simulator time included; Out k and USD: output tokens
  (thousands) and vendor-priced cost per episode (Qwen at the DashScope list price, cache hits at a
  tenth of the input rate).}
  \label{tab:cost-r2r-full}
  \scriptsize
  \setlength{\tabcolsep}{3pt}
  \renewcommand{\arraystretch}{1.0}
  \resizebox{\textwidth}{!}{%
  \begin{tabular}{@{}l r rrrrr r r rr r r rr rr@{}}
    \toprule
    Model & SR & \multicolumn{5}{c}{Turns} & per m & Steps/turn & Out/turn & CoT/turn & In k/turn & s/turn & Out k & In k & Wall s & USD \\
    \cmidrule(lr){3-7}
     &  & all & succ. & fail & budget\,\% & common &  &  &  &  &  &  &  &  &  &  \\
    \midrule
    \texttt{sonnet-5} & 51.3 & 95 & 78 & 113 & 8 & 88 & 9.2 & 2.3 & 114 & n/a & 29.6 & 4.7 & 10.8 & 2805 & 446 & 3.01 \\
    \texttt{opus-4.8} & 55.7 & 66 & 53 & 83 & 3 & 62 & 6.4 & 2.1 & 242 & n/a & 31.8 & 6.6 & 16.1 & 2114 & 437 & 2.59 \\
    \texttt{opus-5} & 70.7 & 53 & 43 & 76 & 0 & 43 & 5.2 & 3.2 & 174 & n/a & 24.1 & 6.3 & 9.2 & 1274 & 332 & 1.27 \\
    \texttt{fable-5} & 68.3 & 63 & 45 & 102 & 3 & 50 & 5.2 & 2.3 & 143 & n/a & 27.4 & 6.0 & 9.0 & 1723 & 379 & 3.43 \\
    \texttt{haiku-5.5} & 49.0 & 50 & 41 & 58 & 12 & 48 & 4.7 & 2.6 & 147 & 78 & 9.6 & 3.1 & 7.3 & 474 & 155 & 0.01 \\
    \texttt{opus-5.5} & 82.3 & 34 & 29 & 57 & 0 & 26 & 3.4 & 3.2 & 85 & 27 & 6.7 & 3.2 & 2.9 & 229 & 108 & 0.21 \\
    \texttt{fable-5.1} & 83.0 & 40 & 35 & 67 & 0 & 30 & 4.0 & 2.8 & 99 & 45 & 13.0 & 5.0 & 4.0 & 524 & 202 & 0.68 \\
    \midrule
    \texttt{gpt-5.5} & 46.0 & 48 & 40 & 56 & 3 & 42 & 4.8 & 2.5 & 60 & 28 & 28.0 & 3.6 & 2.8 & 1308 & 170 & 1.03 \\
    \texttt{gpt-5.6} & 56.7 & 42 & 35 & 51 & 0 & 36 & 4.2 & 3.9 & 85 & 30 & 32.3 & 4.1 & 3.5 & 1346 & 171 & 0.76 \\
    \texttt{gpt-5.6-luna} & 38.0 & 54 & 42 & 62 & 6 & 51 & 5.0 & 3.6 & 91 & 31 & 31.9 & 5.1 & 5.0 & 1733 & 278 & 0.06 \\
    \texttt{gpt-6-luna} & 40.3 & 57 & 47 & 65 & 5 & 50 & 5.6 & 2.7 & 97 & 37 & 36.1 & 4.6 & 5.6 & 2077 & 265 & 0.03 \\
    \texttt{gpt-6-sol} & 69.7 & 49 & 44 & 62 & 2 & 35 & 5.1 & 3.0 & 92 & 40 & 35.3 & 4.4 & 4.5 & 1739 & 218 & 0.47 \\
    \texttt{gpt-6-astra} & 75.7 & 28 & 25 & 36 & 2 & 20 & 3.0 & 3.7 & 55 & 8 & 20.8 & 7.0 & 1.5 & 578 & 193 & 0.90 \\
    \texttt{gpt-6.1-sol} & 75.0 & 30 & 28 & 38 & 1 & 23 & 3.2 & 3.7 & 52 & 7 & 22.0 & 4.6 & 1.6 & 671 & 141 & 0.14 \\
    \midrule
    \texttt{qwen3.5-plus} & 41.3 & 75 & 54 & 90 & 23 & 75 & 6.4 & 2.4 & 78 & n/a & 14.2 & 13.4 & 5.9 & 1068 & 1010 & 0.44 \\
    \texttt{qwen3.6-plus} & 50.0 & 68 & 47 & 88 & 16 & 61 & 5.5 & 2.8 & 90 & n/a & 14.2 & 11.2 & 6.1 & 966 & 757 & 0.50 \\
    \texttt{qwen3.7-plus} & 52.0 & 60 & 44 & 79 & 14 & 58 & 5.2 & 2.9 & 86 & n/a & 13.7 & 10.8 & 5.2 & 831 & 653 & 0.07 \\
    \texttt{qwen3.8-27b} & 45.0 & 73 & 59 & 84 & 15 & 68 & 6.7 & 1.8 & 387 & 346 & 26.5 & 10.8 & 28.0 & 1920 & 781 & 0.34 \\
    \texttt{qwen3.8-max} & 56.3 & 55 & 42 & 73 & 21 & 50 & 5.0 & 2.3 & 293 & n/a & 17.5 & 19.2 & 16.3 & 968 & 1064 & 0.38 \\
    \bottomrule
  \end{tabular}}
\end{table*}

\FloatBarrier

\subsection{Harness configurations}
\label{app:harness}

Harness versions moved over the course of the campaign (the Claude Agent
SDK with its underlying CLI, and the Codex CLI, which updates itself); the
exact version behind every run is recorded in that run's record.

\paragraph{Claude Agent SDK (vendor)~\citep{anthropic2025agentsdk}.}
Subscription authentication. Any ambient API key is stripped so billing
cannot silently change the serving path. The task briefing replaces the
system prompt, removing the product persona. Built-in tools are disabled,
and a strict MCP configuration~\citep{anthropic2024mcp} exposes only our
bridge (\texttt{observe}/\texttt{step} in the minimal-interface
condition). The hard cap is \texttt{max\_turns}=200.

\paragraph{mini-swe-agent (open)~\citep{minisweagent2025}.}
A deliberately small open harness: a plain ReAct loop, a few hundred lines
end to end, on API-key billing, with \emph{no context management}: the
full linear history, including every image, is re-sent on every call, with
prompt caching enabled for Claude models. The briefing is delivered
verbatim, byte-identical to the SDK path's, and the delivered text is
recorded per episode. Hard limits: 200 LLM calls, $2400$\,s wall time.
Model calls are served through LiteLLM~\citep{berriai2026litellm}.

\paragraph{Codex CLI (vendor)~\citep{openai2026codexloop}.}
ChatGPT-subscription authentication. The CLI updates itself, so its
version is recorded per run; the later runs pin the binary and record its
hash. The product persona cannot be removed. The briefing rides as the single user prompt. The CLI runs in a
read-only sandbox with reasoning summaries on and repository-document
injection off. Its built-in shell tool cannot be unmounted, but every use is logged, and
it was invoked in only two of the several thousand logged board episodes. \textbf{Exception to
\cref{tab:frozen}.} The CLI exposes no LLM-call cap, so this harness is
bounded by the action budget and timeout only. Realized call counts stay
far below 200. The plain \texttt{gpt-5.6} and \texttt{gpt-6} identifiers are not served
to ChatGPT-subscription accounts, so the \texttt{gpt-5.6} rows use the
account's \texttt{gpt-5.6-sol} variant and the \texttt{gpt-6} generation
is reported under its served names. Effort tiers are given in
\cref{app:thinking}.

\subsection{Waypoint interface}
\label{app:waypoint}

The waypoint arm of \cref{tab:interface} replaces the primitive action space
with three tools. \texttt{observe()} renders a 12-view RGB-D panorama, feeds
it to a trained candidate-waypoint
predictor~\citep{shi2025smartway}, in the lineage of learned waypoint models
for continuous VLN~\citep{hong2022bridging}, and returns a four-view RGB
strip (left/front/right/back) with up to five candidates drawn as numbered
circles, plus the same options as text. The predictor is an RGB-D
architecture. Our deployment leaves its RGB branch unwired (zeroed), so
prediction is depth-driven, and depth is never shown to the model.
\texttt{goto(k)} sets the agent's heading toward candidate $k$ (a direct
pose write that consumes no simulator steps) and walks its distance through
the same low-level forward primitives, drawing on the same 500-step budget.
\texttt{stop()} ends the episode. Each
episode allows at most 30 \texttt{goto} calls. The 30th ends the episode, so
an agent that exhausts its move budget can no longer issue \texttt{stop()}
and scores zero. Two caveats matter when
reading \cref{tab:interface}. First, the arm changes the observation format
along with the action space (a four-view strip instead of a single front
view), so the deltas measure the combined augmentation. Second, the
predictor is a trained navigation module: the waypoint cells are therefore
not zero-shot in the strict sense of the main board.

\paragraph{The waypoint briefing.}
The waypoint arm's briefing is the primitive briefing of
\cref{app:settings} with only the tool-bound sentences changed, and the
opening user message is the same. Against that text (black), removed words
are struck through and added words are red:

\begin{quote}
\small\ttfamily\raggedright
You are controlling a robot in a real indoor environment (a photorealistic
3D scan of a building). You interact only through these tools:

\bdel{- observe(): look through the robot's forward-facing camera (returns an RGB
image).}\badd{- observe(): look around from where you stand. Returns a panoramic
image (four views labeled Left / Front / Right / Back) with numbered green
circles marking the waypoints you can move to, plus a JSON listing each
waypoint's direction and distance in meters.}

\bdel{- step(actions): execute movement actions in order. 0 = STOP (permanently
ends the episode — declares you have reached the goal), 1 = move forward
0.25 m, 2 = turn left 15 degrees, 3 = turn right 15 degrees.}\badd{- goto(waypoint):
walk to one numbered waypoint from the LATEST observe(). Moving invalidates
the old numbers — observe() again after arriving.}

\badd{- stop(): permanently END the episode, declaring you have reached the goal.}

Your task is to follow this navigation instruction to its endpoint:

"\{instruction\}"

Rules:

- Alternate observing and \bdel{stepping}\badd{moving}: look, decide where the
instruction wants you to go next, move, look again.

- You have a budget of \{budget\} movement actions.

- You succeed only if you \bdel{issue action 0 (STOP)}\badd{call stop()} while
within 3 meters of the instruction's endpoint. \bdel{STOP}\badd{stop()} is
permanent — \bdel{issue}\badd{call} it only when you believe you are at the goal.

\bdel{- Turning in place (e.g. step([2,2,2,2,2,2])) is a cheap way to look around
when unsure.}

- Work autonomously until you stop; nobody can answer questions.
\end{quote}

\subsection{Tools and briefings of every interface arm}
\label{app:interfaces}

Every interface arm is the minimal interface plus a set of tools and the
briefing that names them. The tools are exposed through the same bridge on
every harness; nothing else in the loop changes. This section lists the tools
each arm exposes (\cref{tab:arm-tools}), the description the model sees for
every tool, and every briefing we ran. Variants are shown as word-level
differences against their base text: shared words are black, words removed
from the base are struck through in grey, and words added are red. These
marks are part of the document. Line breaks inside the quoted texts are
normalised for the page; the wording is verbatim. In every briefing
\texttt{\{instruction\}} and \texttt{\{budget\}} are filled per episode
with the instruction, target, or question and the step budget.

\begin{table}[!ht]
  \centering
  
  \caption{Tools exposed by each interface arm. Every arm runs under the frozen budgets of \cref{tab:frozen} (200 model calls, 500 simulator steps, $2400$\,s) unless stated.}
  \label{tab:arm-tools}
  \small
  {%
  \begin{tabular}{@{}lp{0.58\linewidth}l@{}}
    \toprule
    Arm & Tools & Budget notes \\
    \midrule
    Minimal & \texttt{observe}, \texttt{step}; on HM-EQA \texttt{answer}
      replaces \texttt{STOP} & -- \\
    Waypoint & \texttt{observe} (panorama with candidates), \texttt{goto},
      \texttt{stop} & at most 30 \texttt{goto} \\
    Hybrid & \texttt{observe}, \texttt{step}, \texttt{observe\_waypoints},
      \texttt{goto}, \texttt{stop} & at most 30 \texttt{goto} \\
    SLAM & \texttt{observe}, \texttt{step}, \texttt{get\_pose},
      \texttt{get\_map}, \texttt{get\_trajectory}, \texttt{look\_around},
      \texttt{mark}, \texttt{preview\_path} & -- \\
    Hybrid + SLAM & the five Hybrid tools and the six SLAM instruments & at most
      30 \texttt{goto} \\
    Bound & \texttt{step} only; it returns the camera view, and
      \texttt{step([])} looks without moving & 100 model calls \\
    \bottomrule
  \end{tabular}}
\end{table}

\paragraph{Tool descriptions.}
The texts below are the tool descriptions delivered to the model through
the bridge, one per distinct text. Where a tool's description differs by task
family or by arm, the variant is shown as a difference against the base.

\noindent\textbf{\texttt{observe()} (Minimal, Hybrid, SLAM, Hybrid + SLAM).}
\begin{quote}
\small\ttfamily\raggedright

Look through the robot's forward-facing camera. Returns the current egocentric RGB view. Pure read — does not advance the simulator or consume step budget.
\end{quote}

\noindent\textbf{\texttt{step(actions)} (Minimal, Hybrid, SLAM, Hybrid + SLAM; navigation).}
\begin{quote}
\small\ttfamily\raggedright

Execute a sequence of movement actions, in order.

Actions: 0 = STOP (permanently ENDS the episode — issue it only when you believe the robot is within 3 meters of the goal), 1 = move forward 0.25 m, 2 = turn left 15 degrees, 3 = turn right 15 degrees.

Executes sequentially and halts early if the episode ends. Returns how many actions ran, total steps taken, remaining budget, and whether the episode is over. The camera view changes after stepping — call observe() to see the result.
\end{quote}

\noindent\textbf{\texttt{step(actions)}, object-goal variant (against the navigation text).}
\begin{quote}
\small\ttfamily\raggedright
Execute a sequence of movement actions, in order.

Actions: 0 = STOP (permanently ENDS the episode — issue it only when you \bdel{believe }\badd{can see }the \bdel{robot is within 3 meters of the goal), }\badd{target and are standing right next to it), }1 = move forward 0.25 m, 2 = turn left 15 degrees, 3 = turn right 15 \bdel{degrees.}\badd{degrees, 4 = tilt the camera up 30 degrees, 5 = tilt the camera down 30 degrees (tilt changes the camera pitch only, not your position or heading; pitch is clamped to ±60°).}

Executes sequentially and halts early if the episode ends. Returns how many actions ran, total steps taken, remaining budget, and whether the episode is over. The camera view changes after stepping — call observe() to see the result.
\end{quote}

\noindent\textbf{\texttt{step(actions)}, HM-EQA variant (against the navigation text).}
\begin{quote}
\small\ttfamily\raggedright
Execute a sequence of movement actions, in order.

Actions: \bdel{0 = STOP (permanently ENDS the episode — issue it only when you believe the robot is within 3 meters of the goal), }1 = move forward 0.25 m, 2 = turn left \bdel{15 }\badd{30 }degrees, 3 = turn right \bdel{15 degrees.}\badd{30 degrees, 4 = tilt the camera up 30 degrees, 5 = tilt the camera down 30 degrees (tilt changes the camera pitch only, not your position or heading; pitch is clamped to ±60°). There is NO stop action — the episode ends when you call answer().}

Executes sequentially and halts early if the episode ends. Returns how many actions ran, total steps taken, remaining budget, and whether the episode is over. The camera view changes after stepping — call observe() to see the result.
\end{quote}

\noindent\textbf{\texttt{step(actions)}, Bound variant (against the navigation text).}
\begin{quote}
\small\ttfamily\raggedright
Execute a sequence of movement actions, in order.

Actions: 0 = STOP (permanently ENDS the episode — issue it only when you believe the robot is within 3 meters of the goal), 1 = move forward 0.25 m, 2 = turn left 15 degrees, 3 = turn right 15 degrees.

Executes sequentially and halts early if the episode ends. Returns how many actions ran, total steps taken, remaining budget, \bdel{and }whether the episode is \bdel{over. The }\badd{over, and the }camera view \bdel{changes }after \bdel{stepping — call observe() to see }the \bdel{result.}\badd{last action as an attached image. step([]) returns the current view without moving. A finished episode carries no image.}
\end{quote}

\noindent\textbf{\texttt{answer(letter)} (HM-EQA).}
\begin{quote}
\small\ttfamily\raggedright

Permanently END the episode by answering the multiple-choice question. Pass a single letter: "A", "B", "C" or "D". This is your one and only answer — call it once you have seen enough of the building to be confident. An episode that ends without answer() scores zero.
\end{quote}

\noindent\textbf{\texttt{observe()} on the Waypoint arm, identical to \texttt{observe\_waypoints()} on Hybrid and Hybrid + SLAM.}
\begin{quote}
\small\ttfamily\raggedright

Look around from where you stand. Returns a panoramic image of four views labeled Left / Front / Right / Back with numbered green circles marking the waypoints you can move to, plus a JSON listing each waypoint's direction, angle (degrees left of your heading; negative = right) and distance in meters. Pure read — does not advance the simulator or consume step budget.
\end{quote}

\noindent\textbf{\texttt{goto(k)} (Waypoint).}
\begin{quote}
\small\ttfamily\raggedright

Move to one numbered waypoint from the LATEST observe() result: the robot turns toward it and walks there. Moving invalidates the old numbers — call observe() again afterwards to see the new surroundings and waypoints. Returns steps consumed, remaining budget, and whether the episode is over.
\end{quote}

\noindent\textbf{\texttt{goto(k)} on Hybrid and Hybrid + SLAM (against the Waypoint text).}
\begin{quote}
\small\ttfamily\raggedright
Move to one numbered waypoint from the LATEST \bdel{observe() }\badd{observe\_waypoints() }result: the robot turns toward it and walks there. Moving invalidates the old numbers — call \bdel{observe() }\badd{observe\_waypoints() }again afterwards to see the new surroundings and waypoints. Returns steps consumed, remaining budget, and whether the episode is over.
\end{quote}

\noindent\textbf{\texttt{stop()} (Waypoint, Hybrid, Hybrid + SLAM; navigation).}
\begin{quote}
\small\ttfamily\raggedright

Permanently END the episode, declaring you have reached the goal. Issue it only when you believe the robot is within 3 meters of the instruction's endpoint — stopping is irreversible.
\end{quote}

\noindent\textbf{\texttt{stop()}, object-goal variant (against the navigation text).}
\begin{quote}
\small\ttfamily\raggedright
Permanently END the episode, declaring you have \bdel{reached }\badd{found }the \bdel{goal. }\badd{target. }Issue it only when you \bdel{believe }\badd{can see }the \bdel{robot is within 3 meters of the instruction's endpoint }\badd{target and are standing right next to it }— stopping is irreversible.
\end{quote}

\noindent\textbf{\texttt{get\_pose()} (SLAM, Hybrid + SLAM).}
\begin{quote}
\small\ttfamily\raggedright

Read the robot's SLAM-estimated pose: position (x, z) in meters and heading yaw\_deg, in a fixed frame anchored at your start pose (x = right and z = forward OF YOUR STARTING POSE; yaw\_deg is 0 at your starting heading and INCREASES when you turn right). The same x/z coordinates label the gridlines on get\_map(), so pose numbers place you on the map. Free — no step cost. slam="slam\_lost" means the estimate is currently stale.
\end{quote}

\noindent\textbf{\texttt{get\_map()} (SLAM, Hybrid + SLAM).}
\begin{quote}
\small\ttfamily\raggedright

Read the occupancy map built automatically as you move

(it already contains an initial 360-degree scan of your start location).

Returns a top-down image cropped to what you have explored — UP on the image

is your STARTING heading (+z), right is +x. On it: white = walked-free

space, black = obstacle, gray = unexplored; the blue ARROW is you, pointing

your current heading; the blue LINE is your path so far; the "S" circle is

your start; orange dots are markers you dropped with mark(). Gridlines

every 2 m are labeled on both axes with the same x/z coordinates get\_pose()

reports (the 0-axes are drawn heavier), and the crop window never pans or

rescales between calls — it only grows in 2 m steps as you explore, so

coordinates stay put from map to map. By default the map shows YOUR CURRENT

FLOOR; a "floor k/n" tag (top right) says which storey you are on and how

many have been seen. Pass floor=k (the tag's 1-based number) to view

another storey's layer, or zoom=true to get TWO images — the full map plus

a fixed 8 m close-up centered on you (current floor, same coordinate

lattice) — useful when the full map has grown too coarse for nearby

detail. Free — no step cost. SLAM estimates, not ground truth.
\end{quote}

\noindent\textbf{\texttt{get\_trajectory()} (SLAM, Hybrid + SLAM).}
\begin{quote}
\small\ttfamily\raggedright

Read the path you have walked ON THIS TASK as (x, z) points in the same fixed frame as get\_pose(), oldest first. Useful to check whether you are circling or which areas you already covered since this goal began. The map keeps the whole visit; this is just the current stretch. Free — no step cost.
\end{quote}

\noindent\textbf{\texttt{look\_around()} (SLAM, Hybrid + SLAM).}
\begin{quote}
\small\ttfamily\raggedright

Turn a full circle in place and return four camera views 90 degrees apart — front, right, back, left relative to your CURRENT heading — ending faced exactly where you started. The camera's field of view is 90 degrees, so together the four views tile the complete panorama. This is a pure LOOK: like observe(), it does not touch SLAM or the map — get\_map() stays exactly as it was, and only walking builds the map. Free — no step cost, no budget; turning in place cannot collide.
\end{quote}

\noindent\textbf{\texttt{mark(label)} (SLAM, Hybrid + SLAM).}
\begin{quote}
\small\ttfamily\raggedright

Drop a named marker at your current position — it appears as an orange dot with your label on every later get\_map() (on the floor where you dropped it), and in the map's JSON. Use it to remember places the instruction cares about ("kitchen", "blue door") instead of trusting visual memory. Re-marking an existing label MOVES it to where you are now. Labels cap at 24 characters. Free — no step cost.
\end{quote}

\noindent\textbf{\texttt{preview\_path(x, z)} (SLAM, Hybrid + SLAM).}
\begin{quote}
\small\ttfamily\raggedright

Preview a walkable route from you to map point (x, z) — the same x/z coordinates get\_pose() and the map gridlines use. Plans over the walked-free (white) cells of your current floor's map and returns the map image with the route drawn in magenta (X = target) plus the route length in meters. If (x, z) is not on known free space it snaps to the nearest free cell within 1 m, or reports it unreachable — the map only knows space you have already seen. NOTHING MOVES: this is a read-only preview. Free — no step cost.
\end{quote}

\noindent\textbf{\texttt{mark(label)}, object-goal variant (against the navigation text).}
\begin{quote}
\small\ttfamily\raggedright
Drop a named marker at your current position — it appears as an orange dot with your label on every later get\_map() (on the floor where you dropped it), and in the map's JSON. Use it to remember places \bdel{the instruction cares about }\badd{that matter to your search }("kitchen", \bdel{"blue door") }\badd{"bedroom door", "unchecked hallway") }instead of trusting visual memory. Re-marking an existing label MOVES it to where you are now. Labels cap at 24 characters. Free — no step cost.
\end{quote}

\paragraph{Briefings.}
The base text is the minimal navigation briefing reproduced in
\cref{app:frozen}; the Waypoint briefing is given against it in
\cref{app:waypoint}. Every other briefing we ran follows, each as a
difference against its base. The opening user message is ``Begin navigating. Call observe() first to see where you are.'' for every navigation and object-goal
briefing, ``Begin exploring. Call observe() first to see where you are.'' for HM-EQA, and
``Begin navigating. Call step([]) first to see where you are.'' for the Bound arm.

\noindent\textbf{Minimal, object-goal navigation (HM3D-ObjectNav v1 and v2, MP3D-ObjectNav), against the navigation briefing.}
\begin{quote}
\small\ttfamily\raggedright
You are controlling a robot in a real indoor environment (a photorealistic 3D scan of a building). You interact only through these tools:

- observe(): look through the robot's forward-facing camera (returns an RGB image).

- step(actions): execute movement actions in order. 0 = STOP (permanently ends the episode — declares you have \bdel{reached }\badd{found }the \bdel{goal), }\badd{target), }1 = move forward 0.25 m, 2 = turn left 15 degrees, 3 = turn right 15 \bdel{degrees.}\badd{degrees, 4 = tilt the camera up 30 degrees, 5 = tilt the camera down 30 degrees (tilt changes the camera pitch only, not your position or heading).}

Your task is \badd{object-goal navigation: no route is given — search the building until you find the target object, walk up }to \bdel{follow this navigation instruction to its endpoint:}\badd{it, and stop there.}

\badd{Target object: }"\{instruction\}"

Rules:

- Alternate observing and stepping: look, decide where \bdel{the instruction wants you }to \bdel{go }\badd{search }next, move, look again.

\badd{- Explore efficiently: sweep toward where a "\{instruction\}" is most likely to be, and avoid re-walking areas you have already ruled out.}

- You have a budget of \{budget\} movement actions.

- You succeed only if you issue action 0 (STOP) while \bdel{within 3 meters of the instruction's endpoint. }\badd{standing right next to a "\{instruction\}" (within about 1 meter). Any instance counts. }STOP is permanent — issue it only when you \bdel{believe you }\badd{can see the target and }are \bdel{at the goal.}\badd{standing right next to it.}

- Turning in place (e.g. \bdel{step([2,2,2,2,2,2])) }\badd{step([2,2,2])) }is a cheap way to look around when unsure.

\badd{- Tilt changes persist until you change them again; level the camera again before moving on, and check your camera pitch before interpreting a view.}

- Work autonomously until you stop; nobody can answer questions.
\end{quote}

\noindent\textbf{Minimal, HM-EQA, against the navigation briefing.}
\begin{quote}
\small\ttfamily\raggedright
You are controlling a robot in a real indoor environment (a photorealistic 3D scan of a building). \badd{Its camera starts tilted 30 degrees downward. }You interact only through these tools:

- observe(): look through the robot's forward-facing camera (returns an RGB image).

- step(actions): execute movement actions in order. \bdel{0 = STOP (permanently ends the episode — declares you have reached the goal), }1 = move forward 0.25 m, 2 = turn left \bdel{15 }\badd{30 }degrees, 3 = turn right \bdel{15 degrees.}\badd{30 degrees, 4 = tilt the camera up 30 degrees, 5 = tilt the camera down 30 degrees (tilt changes the camera pitch only, not your position or heading). There is no stop action.}

\badd{- answer(letter): permanently END the episode by answering the question with "A", "B", "C" or "D".}

Your task is \bdel{to follow }\badd{embodied question answering: explore the building until you can answer }this \bdel{navigation instruction to its endpoint:}\badd{multiple-choice question about it:}

"\{instruction\}"

Rules:

- Alternate observing and stepping: look, decide where \bdel{the instruction wants you }to go \bdel{next, }\badd{to find the evidence the question needs, }move, look again.

- You have a budget of \{budget\} movement \bdel{actions.}\badd{actions. If it runs out you can still observe and answer from where you stand.}

- You succeed only if \bdel{you issue action 0 (STOP) while within 3 meters of }\badd{answer() gives }the \bdel{instruction's endpoint. STOP }\badd{correct letter. answer() }is permanent — \bdel{issue }\badd{call }it \bdel{only when }\badd{once }you \bdel{believe you are at the goal.}\badd{have seen enough evidence to be confident, and always answer before ending: an episode without answer() scores zero.}

- Turning in place (e.g. \bdel{step([2,2,2,2,2,2])) }\badd{step([2,2,2])) }is a cheap way to look around when unsure.

\badd{- Tilt changes persist until you change them again; level the camera again before moving on, and check your camera pitch before interpreting a view.}

- Work autonomously until you \bdel{stop; }\badd{answer; }nobody can \bdel{answer questions.}\badd{help you.}
\end{quote}

\noindent\textbf{Hybrid, against the navigation briefing.}
\begin{quote}
\small\ttfamily\raggedright
You are controlling a robot in a real indoor environment (a photorealistic 3D scan of a building). You interact only through these tools:

- observe(): look through the robot's forward-facing camera (returns an RGB image).

- step(actions): execute movement actions in order. 0 = STOP (permanently ends the episode — declares you have reached the goal), 1 = move forward 0.25 m, 2 = turn left 15 degrees, 3 = turn right 15 degrees.

\badd{- observe\_waypoints(): look around from where you stand. Returns a panoramic image (four views labeled Left / Front / Right / Back) with numbered green circles marking the waypoints you can move to, plus a JSON listing each waypoint's direction and distance in meters.}

\badd{- goto(waypoint): walk to one numbered waypoint from the LATEST observe\_waypoints(). Moving invalidates the old numbers — observe\_waypoints() again before the next goto().}

\badd{- stop(): permanently END the episode, declaring you have reached the goal.}

Your task is to follow this navigation instruction to its endpoint:

"\{instruction\}"

Rules:

- Alternate observing and \bdel{stepping: }\badd{moving: }look, decide where the instruction wants you to go next, move, look again.

- You have a budget of \{budget\} movement actions.

\badd{- You may make at most 30 waypoint moves; each observe\_waypoints() and goto() result reports how many remain. When they run out the episode ends, so do not wander.}

- You succeed only if you issue action 0 (STOP) while within 3 meters of the instruction's endpoint. STOP is permanent — issue it only when you believe you are at the goal.

- Turning in place (e.g. step([2,2,2,2,2,2])) is a cheap way to look around when unsure.

- Work autonomously until you stop; nobody can answer questions.
\end{quote}

\noindent\textbf{SLAM, against the navigation briefing (the instrument block is appended unchanged).}
\begin{quote}
\small\ttfamily\raggedright
You are controlling a robot in a real indoor environment (a photorealistic 3D scan of a building). You interact only through these tools:

- observe(): look through the robot's forward-facing camera (returns an RGB image).

- step(actions): execute movement actions in order. 0 = STOP (permanently ends the episode — declares you have reached the goal), 1 = move forward 0.25 m, 2 = turn left 15 degrees, 3 = turn right 15 degrees.

Your task is to follow this navigation instruction to its endpoint:

"\{instruction\}"

Rules:

- Alternate observing and stepping: look, decide where the instruction wants you to go next, move, look again.

- You have a budget of \{budget\} movement actions.

- You succeed only if you issue action 0 (STOP) while within 3 meters of the instruction's endpoint. STOP is permanent — issue it only when you believe you are at the goal.

- Turning in place (e.g. step([2,2,2,2,2,2])) is a cheap way to look around when unsure.

- Work autonomously until you stop; nobody can answer questions.

\badd{You additionally have six FREE instrument tools backed by an onboard SLAM}

\badd{system (no step cost, no budget, callable any time):}

\badd{- get\_pose(): your estimated position (x, z, meters) and heading (yaw\_deg) in a fixed frame anchored at your start pose.}

\badd{- get\_map(): a top-down occupancy map built automatically as you move (it already contains an initial 360-degree scan of your start location). Its coordinate grid stays fixed between calls, and it shows your current floor's layer — a "floor k/n" tag marks the storey when a scene has several; pass floor=k to view another storey, or zoom=true to get both the full map and a fixed 8-meter close-up around you. Markers you drop with mark() appear on it.}

\badd{- get\_trajectory(): the path you have walked on this task, oldest first (the map keeps the whole visit).}

\badd{- look\_around(): turn a full circle in place and get four views 90 degrees apart (front / right / back / left of your current heading), ending faced where you started. A pure look, like observe(): it does not change the map — only walking builds the map.}

\badd{- mark(label): drop a named marker at your position (re-marking a label moves it) — it shows on the map, so you can pin places the instruction mentions instead of memorizing them.}

\badd{- preview\_path(x, z): a read-only route preview over the map's walked-free space to point (x, z), drawn on the map with its length in meters — nothing moves.}
\end{quote}

\noindent\textbf{Hybrid + SLAM, navigation.} The briefing is the Hybrid briefing followed by the instrument block of the SLAM briefing, byte for byte; it is not repeated here.

\noindent\textbf{Hybrid + SLAM, object-goal navigation, against the object-goal Minimal briefing.}
\begin{quote}
\small\ttfamily\raggedright
You are controlling a robot in a real indoor environment (a photorealistic 3D scan of a building). You interact only through these tools:

- observe(): look through the robot's forward-facing camera (returns an RGB image).

- step(actions): execute movement actions in order. 0 = STOP (permanently ends the episode — declares you have found the target), 1 = move forward 0.25 m, 2 = turn left 15 degrees, 3 = turn right 15 degrees, 4 = tilt the camera up 30 degrees, 5 = tilt the camera down 30 degrees (tilt changes the camera pitch only, not your position or heading).

\badd{- observe\_waypoints(): look around from where you stand. Returns a panoramic image (four views labeled Left / Front / Right / Back) with numbered green circles marking the waypoints you can move to, plus a JSON listing each waypoint's direction and distance in meters.}

\badd{- goto(waypoint): walk to one numbered waypoint from the LATEST observe\_waypoints(). Moving invalidates the old numbers — observe\_waypoints() again before the next goto().}

\badd{- stop(): permanently END the episode, declaring you have found the target.}

Your task is object-goal navigation: no route is given — search the building until you find the target object, walk up to it, and stop there.

Target object: "\{instruction\}"

Rules:

- Alternate observing and stepping: look, decide where to search next, move, look again.

- Explore efficiently: sweep toward where a "\{instruction\}" is most likely to be, and avoid re-walking areas you have already ruled out.

- You have a budget of \{budget\} movement actions.

\badd{- You may make at most 30 waypoint moves; each observe\_waypoints() and goto() result reports how many remain. When they run out the episode ends, so do not wander.}

- You succeed only if you issue action 0 (STOP) while standing right next to a "\{instruction\}" (within about 1 meter). Any instance counts. STOP is permanent — issue it only when you can see the target and are standing right next to it.

- Turning in place (e.g. step([2,2,2])) is a cheap way to look around when unsure.

- Tilt changes persist until you change them again; level the camera again before moving on, and check your camera pitch before interpreting a view.

- Work autonomously until you stop; nobody can answer questions.

\badd{You additionally have six FREE instrument tools backed by an onboard SLAM}

\badd{system (no step cost, no budget, callable any time):}

\badd{- get\_pose(): your estimated position (x, z, meters) and heading (yaw\_deg) in a fixed frame anchored at your start pose.}

\badd{- get\_map(): a top-down occupancy map built automatically as you move (it already contains an initial 360-degree scan of your start location). Its coordinate grid stays fixed between calls, and it shows your current floor's layer — a "floor k/n" tag marks the storey when a building has several; pass floor=k to view another storey, or zoom=true to get both the full map and a fixed 8-meter close-up around you. Markers you drop with mark() appear on it.}

\badd{- get\_trajectory(): the path you have walked on this task, oldest first (the map keeps the whole visit).}

\badd{- look\_around(): turn a full circle in place and get four views 90 degrees apart (front / right / back / left of your current heading), ending faced where you started. A pure look, like observe(): it does not change the map — only walking builds the map.}

\badd{- mark(label): drop a named marker at your position (re-marking a label moves it) — it shows on the map, so you can pin rooms you have checked and spots worth returning to instead of memorizing them.}

\badd{- preview\_path(x, z): a read-only route preview over the map's walked-free space to point (x, z), drawn on the map with its length in meters — nothing moves.}
\end{quote}

\noindent\textbf{Bound, against the navigation briefing.}
\begin{quote}
\small\ttfamily\raggedright
You are controlling a robot in a real indoor environment (a photorealistic 3D scan of a building). You interact only through these tools:

\bdel{- observe(): look through the robot's forward-facing camera (returns an RGB image).}

- step(actions): execute movement actions in \bdel{order. }\badd{order and return the camera view after the last one; step([]) just looks without moving. }0 = STOP (permanently ends the episode — declares you have reached the goal), 1 = move forward 0.25 m, 2 = turn left 15 degrees, 3 = turn right 15 degrees.

Your task is to follow this navigation instruction to its endpoint:

"\{instruction\}"

Rules:

\bdel{- Alternate observing and stepping: look, }\badd{- Every step() result shows what the camera sees after your actions: }decide where the instruction wants you to go next, move, \bdel{look again.}\badd{read the new view.}

- You have a budget of \{budget\} movement actions.

- You succeed only if you issue action 0 (STOP) while within 3 meters of the instruction's endpoint. STOP is permanent — issue it only when you believe you are at the goal.

- Turning in place (e.g. step([2,2,2,2,2,2])) is a cheap way to look around when unsure.

- Work autonomously until you stop; nobody can answer questions.
\end{quote}

\end{document}